\documentclass[10pt,journal]{IEEEtran}
\ifCLASSOPTIONcompsoc
\usepackage[nocompress]{cite}
\else
\usepackage{cite}
\fi
\usepackage{graphicx}
\usepackage{multirow}
\ifCLASSINFOpdf
\else
\fi
\usepackage{times}
\usepackage{epsfig}
\usepackage{graphicx}
\usepackage{amsmath}
\usepackage{amssymb}
\usepackage{amsfonts}
\usepackage{mathrsfs}
\usepackage{color}
\usepackage{physics}
\usepackage{subfig}

\usepackage{float}

\newcommand{\best}[1]{{\textbf{#1}}}

\newcommand{\eg}{{\textit{e.g., }}}
\newcommand{\ie}{{\textit{i.e., }}}

\graphicspath{{figs/},{authors/},{./}}

\begin{document}
	
	%
	
	\title{Self-supervised Re-renderable Facial Albedo Reconstruction from Single Image}

\author{
    Mingxin Yang,~
    Jianwei Guo,~
    Zhanglin Cheng,~
    Xiaopeng Zhang,~
    Dong-Ming Yan
    \IEEEcompsocitemizethanks{
    \IEEEcompsocthanksitem M. Yang and Z. Cheng are with Shenzhen Institute of Advanced Technology (SIAT), Chinese Academy of Sciences, Shenzhen 518055, China.
     \IEEEcompsocthanksitem J. Guo, X. Zhang and D.M. Yan are with National Laboratory of Pattern Recognition (NLPR), Institute of Automation, Chinese Academy of Sciences, Beijing 100190, China.
     }
	\thanks{}
}
	
	%
	%

	\markboth{Journal of \LaTeX\ Class Files,~Vol.~pp, No.~99, 2022}%
	{Shell \MakeLowercase{\textit{et al.}}: Bare Demo of IEEEtran.cls for Computer Society Journals}
	%



	\IEEEtitleabstractindextext{%
		\begin{abstract}
	Reconstructing high-fidelity 3D facial texture from a single image is a quite challenging task due to the lack of complete face information and the domain gap between the 3D face and 2D image.  
	Further, obtaining re-renderable 3D faces has become a strongly desired property in many applications, where the term 're-renderable' demands the facial texture to be spatially complete and disentangled with environmental illumination.
	In this paper, we propose a new self-supervised deep learning framework for reconstructing high-quality and re-renderable facial albedos from single-view images in-the-wild.
   Our main idea is to first utilize a \emph{prior generation module} based on the 3DMM proxy model to produce an unwrapped texture and a globally parameterized prior albedo. Then we apply a \emph{detail refinement module} to synthesize the final texture with both high-frequency details and completeness.  
    To further make facial textures disentangled with illumination, we propose a novel detailed illumination representation which is reconstructed with the detailed albedo together.
    We also design several novel regularization losses on both the albedo and illumination maps to facilitate the disentanglement of these two factors. 
    Finally, by leveraging a differentiable renderer, each face attribute can be jointly trained in a self-supervised manner without requiring ground-truth facial reflectance.
    Extensive comparisons and ablation studies on challenging datasets demonstrate that our framework outperforms state-of-the-art approaches. 

	\end{abstract}
		
		\begin{IEEEkeywords}
			facial albedo reconstruction, 3D face, self-supervised training, illumination reconstruction
	\end{IEEEkeywords}}
	
	\maketitle

	\IEEEdisplaynontitleabstractindextext

	%
	\IEEEpeerreviewmaketitle

\section{Introduction}\label{sec:introduction}
Reconstructing high-fidelity 3D human faces is a longstanding problem in multimedia and computer vision communities. 
This task aims to estimate a realistic 3D facial representation, \ie predicting face geometry, appearance, expression, and scene lighting from the input source. Faithfully reconstructing 3D faces is a crucial prerequisite for many downstream applications including face editing~\cite{thies2016face2face}, virtual avatar generation~\cite{Chen20223D,lattas2020avatarme}, face alignment~\cite{Tu20213D} and recognition~\cite{Li2017Multimodal}. 

\begin{figure}[!t]
\centering
  \includegraphics[width=0.95\linewidth]{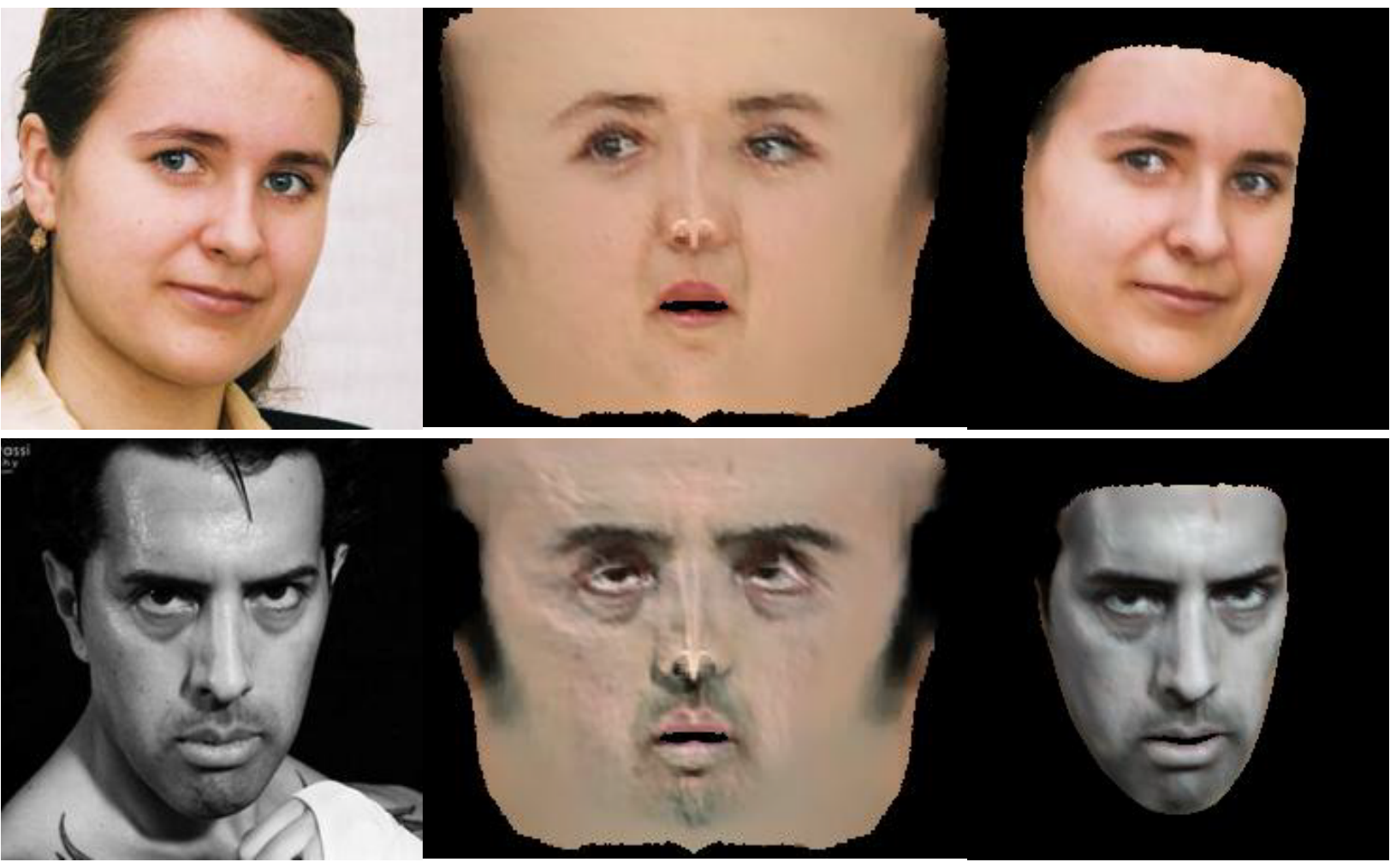}   
  \caption{We learn to reconstruct high-fidelity facial textures from in-the-wild images. Left: input single image; Middle: detailed albedo map generated by our neural network; Right: re-rendered results using our detailed albedo and fine-grained illumination maps.}
  \label{fig:show}
\end{figure}

Recently, single image based 3D face reconstruction has gained much attention. 
However, it is a highly challenging and ill-posed problem due to the domain gap between the 3D face and 2D image. 
To learn the mapping from a single image to a 3D face, a parametric model called \emph{3D Morphable Model} (3DMM)~\cite{Blanz2002A} is developed as the prior model of a 3D human face that transforms the 3D reconstruction to a parameter estimation problem. 
However, 3DMM largely limits the representation capability of the parametric model because it was constructed by applying linear subspace modelling techniques on a limited number of 3D face scans, thus leading to poor reconstruction fidelity when being applied to in-the-wild images~\cite{deng2019accurate,genova2018unsupervised,tewari2017mofa}.
Recently, many attempts have been conducted to tackle the detail lacking drawback of 3DMM by adding non-linearity into the parametric model, for example, replacing the linear 3DMM with a completely non-linear one~\cite{gecer2019ganfit,tran2019towards,tran2018nonlinear} or complementing non-linearity upon the 3DMM coarse reconstruction~\cite{jiangke2020towards,chen2019self,chen2019photo,tewari2019fml, tewari2018self,jackson2017large}. In these methods, facial details are either represented in geometry by a displacement map or encoded into appearance by a detailed texture map (or albedo map). In this work, we focus on high-fidelity appearance reconstruction and apply a coarse-to-fine approach to generate textures that capture many facial details.


The methods for reconstructing high-fidelity facial textures can be further roughly divided into two categories. 
The first category extends the basic idea of the parametric model and utilizes a self-collected facial texture dataset to train a generative model~\cite{genova2018unsupervised,gecer2019ganfit,lattas2020avatarme, lattas2021avatarme++}. 
When estimating a new image, these approaches fit the closest texture in the subspace to the input. They could achieve high-quality results even when the inputs are occluded or in extreme light conditions. 
However, their generation results can not maintain the idiosyncrasy of the human faces well because of the limited representation capacity of the generative model.
The other category of methods typically reconstructs the texture directly from the input image~\cite{jiangke2020towards,chen2019photo}. 
Although their reconstruction corresponds to input image better, their reconstruction quality is highly influenced by the input image, and noise-like occlusion and extreme environmental illumination will cause artifacts in the reconstructed texture.

    
Apart from the requirements of high-fidelity texture reconstruction, many applications (\eg virtual avatar generation) demand the texture to be re-renderable. Specifically, the reconstructed texture should be not only faithful to the input image, but also disentangled with illumination (which is referred to as an albedo). 
However, above mentioned methods can not solve the disentanglement of face albedo with illumination. 
The reasons are two-fold: (1) real facial textures are difficult to capture without illumination.~\cite{gecer2019ganfit}; and (2) the widely used three-band \emph{spherical harmonics} (SH) lighting model has a limited representational capacity~\cite{jiangke2020towards}. 

To address these mentioned issues, we propose a new self-supervised learning algorithm that takes both the advantages of above two categories of methods to generate high-fidelity and re-renderable facial albedos.
Our method adopts a coarse-to-fine paradigm which first utilizes a \emph{prior albedo generation module} to produce a coarse albedo as a prior, then adds facial details on the prior albedo by using a \emph{detail refinement module}. 
Specifically, we adopt a pre-trained inference network based on 3DMM to produce a prior albedo from the input image. 
Then, we transform the prior albedo into a complete and detailed facial texture by employing an image-to-image translation network to preserve high-frequency details.
In addition, we introduce a novel detailed illumination representation and propose a detailed illumination decoder to make the albedo disentangled with environmental illumination. This property is especially useful for rendering from novel view points.
Several regularization loss functions are designed on both the illumination side and albedo side for achieving a high-fidelity and re-renderable albedo.
Finally, our pipeline can be efficiently trained in a self-supervised manner with the help of differentiable rendering~\cite{genova2018unsupervised}. Fig.~\ref{fig:show} gives two examples of our reconstruction results. 
In summary, our work makes the following contributions:
\begin{itemize}
\item We propose a new self-supervised neural network to obtain a high-fidelity and re-renderable facial albedo. We are able to deal with potential occlusions commonly existed in facial images. The self-supervised learning further makes our approach generalize well among other unseen data in-the-wild.
\item We devise a novel representation of detailed illumination by a localized spherical harmonics to achieve a more accurate illumination estimation, which alleviates the limited expressiveness of widely used SH-based lighting model.
\item We design several regularization losses to ensure that the detailed albedo is similar to the prior coarse albedo while keeping high-frequency details. Especially, the cross perceptual loss is effective to disentangle lighting from person-specific details such as beards and wrinkles.
\end{itemize} 

\section{Related work}
\label{sec:related}

\subsection{Parametric Models for the Human Face}
\label{subsec:FPM}
The seminal parametric model of 3DMM was first introduced by Blanz and Vetter~\cite{Blanz2002A}. This work applies a subspace modeling method on a collected 3D face scans and produces low-dimensional representations for facial identity, expression and albedo. Since the 3DMM was proposed, many variants~\cite{cao2013facewarehouse,Ferrari2017,li2017learning,booth2018large} have extended it to obtain better performance.  
Egger et al.~\cite{3DMM_survey} provide a comprehensive survey of 3DMM. 
To improve representation power, parametric models with non-linearity are introduced~\cite{FLAME,tran2019towards}. 
Although this model expands the representation capacity of 3DMM, the local modeling scheme leads to stitching artifacts in the generated result. 
Ganfit~\cite{gecer2019ganfit} utilizes a progressive GAN \cite{karras2017progressive} to construct a generative parametric model, which collects a dataset of high-resolution human facial textures and trains the network on it. However, the model has the drawback that the illumination is baked into the texture. Lattas et al.~\cite{lattas2020avatarme} extend Ganfit by post-processing (super-resolution, de-lighting and BRDF inference) the derived texture. 
Its limitation is that the captured dataset does not contain sufficient samples on different ethnicities and may produce unfaithful results.


\subsection{3D Face Reconstruction}
\noindent\textbf{Monocular face reconstruction. }
Zollhoefer et al.~\cite{zollhofer2018state} gives a state-of-the-art report summarizes recent trends in monocular facial reconstruction, tracking, and applications.
Given the lack of depth information in RGB images, 3DMM is always included as a proxy model in monocular face reconstruction pipeline. A variety of works~\cite{chen2019photo,gecer2019ganfit,thies2016face2face,deng2019accurate,genova2018unsupervised,tewari2018self,tewari2017mofa,richardson2017learning,Fan2021Dual} utilize this paradigm to transform the reconstruction to a parameter estimation problem.  
These works can be further divided into two categories by the inferring approaches: fitting-based and learning-based methods. The former provides more accurate reconstruction results but consumes more time, while the latter leverages deep convolutional neural networks to estimate 3DMM parameters leading to a fast inference. 

\noindent\textbf{High-fidelity 3D face reconstruction.} 
Although 3DMM is capable of reconstructing 3D face roughly, it lacks detailed information. Accordingly, the reconstruction results will lack certain characteristics such as wrinkles and pores. Recently, many methods were proposed for high-fidelity 3D reconstruction.
The direct idea is to capture a dataset with high-quality 3D face ground-truth and train the inference deep network on the dataset~\cite{yang2020facescape,chen2019photo,nagano2018pagan,yamaguchi2018high-fidelity,GZCVVPT16} to achieve authentic reconstruction results. However, constructing such datasets requires expensive capture equipment (\eg LED sphere) and leads to laborious work. Meanwhile, the data are mostly captured in a controlled environment and the network trained on it is hard to handle in-the-wild face images.

\begin{figure*}[!thb]
\centering
  \includegraphics[width=1.0\linewidth]{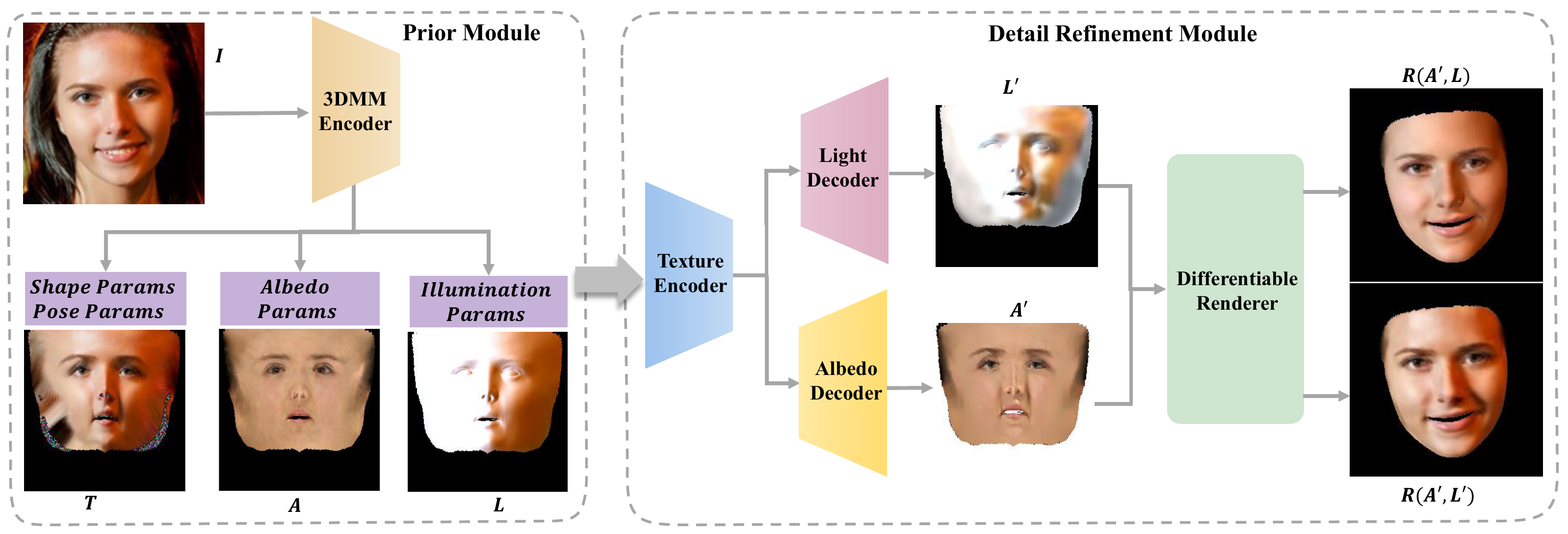}   
  \caption{\textbf{Network architecture for 3D face reconstruction:} The left part shows our prior module that takes a 3DMM as proxy model and generates unwrapped facial texture map ($T$), prior albedo map ($A$) and prior illumination ($L$). Our prior module includes a pre-trained 3DMM encoder to regress 3DMM parameters from input image ($I$) and a fixed 3DMM linear decoder to generate corresponding attributes of the 3D face. The unwrapped facial texture map and prior albedo map are then fed into our detail refinement module (right part in the figure) to generate the detailed albedo ($A^{'}$) and detailed illumination ($L^{'}$). The detail refinement module contains one texture encoder with a light decoder and an albedo decoder. After getting the detailed illumination map and the detailed albedo map, they are sent to a differentiable renderer with other 3D face attributes to obtain re-rendered images ($R(A^{'},L^{'}),R(A^{'},L)$) for self-supervision. }
  \label{fig:overview}
\end{figure*}

With the development of CNNs and the introduction of differentiable rendering, a self-supervised paradigm with re-rendering loss is incorporated into the facial detail reconstruction~\cite{jiangke2020towards, tewari2018self,tewari2019fml,chen2019self,tran2018nonlinear,tran2019towards}. These works add a regression network to complement detail information upon the 3DMM coarse reconstruction. However, the coarse model and detail model are often trained separately. These approaches are trained in in-the-wild image datasets and resolve the drawbacks of supervised approaches. Wu et al.~\cite{shangzhe2019unsupervised} leverage the symmetry in the human face and directly regress the depth maps; however, it is unable to reconstruct a complete face model and produce artifacts when encountering extreme inputs (e.g., images with non-frontal faces).
Other than the above-mentioned 3D face reconstruction methods, another branch of work focuses on human portrait video synthesis. Many of these methods~\cite{gafni2021nerface,deferred19,kim2018deep} still utilize the 3DMM as an intermediate representation and generate high-fidelity details upon it. However, these works do not produce any detailed 3D face model;hence, their reconstruction results can not be directly utilized by graphic renderers.

\noindent\textbf{Representation of facial details. }
3D facial detail information can be modeled in either geometric space as displacements or normal~\cite{decasig2021, tran2019towards,chen2019photo,tran2018nonlinear,xueying2020lightweight} or in appearance space as texture or albedo~\cite{jiangke2020towards,gecer2019ganfit,deng2018uv-gan}, or both of them~\cite{Chai2021Expression}.
Considering the representation space, details can be represented as maps in uv-space, or maps in frontal-face space, or vertex attributes on a 3D face mesh.
The methods of~\cite{chen2019photo,tran2019towards} represent facial detail in uv-space. Chen et al.~\cite{chen2019photo} unwrap partial image texture from image and regress a detailed displacement map from it; Tran et al.~\cite{tran2019towards} directly generate a uv representation of texture and geometry from the input image. 
Lin et al.~\cite{jiangke2020towards} utilize graph convolutional network (GCN) and model texture as three-channel vertex attributes on a 3D face mesh to obtain competitive results. Wu et al.~\cite{shangzhe2019unsupervised} take advantage of the symmetry characteristic of human faces and represent the depth, albedo and illumination maps in the frontal-face space. Although applying GCN on mesh could produce convincing results, we argue that uv-representation is still a valid representation for detail reconstruction because the proper face parameterization keeps most of the face topology and can be easily processed by 2D CNN. 
In this work, we present a monocular high-fidelity 3D face reconstruction approach and represent the detail information by a detailed texture map in the uv-space.




\subsection{Image Formation Modeling}
Image formation is the process that maps a 3D model with an environmental condition to a 2D image space. The core in the process is the reflectance models that include illumination modeling and interaction pattern between light and the model surface. In 3D face modeling, three-band RGB spherical harmonic lighting representation~\cite{zhang2006face} and Lambertian surface are often considered the default settings~\cite{jiangke2020towards,deng2019accurate,tran2019towards}. Spherical harmonics is a set of orthonormal basis defined on a sphere that is analogous to Fourier basis in the Euclidean space, and it can be a proper approximation to illumination in the natural lighting. Lambertian surface assumes that the surface irradiance is irrelevant to the observer's position and only depends on the incident light direction. Although these two assumptions provide proper approximation, they neglect other reflection effects (such as specular reflection) in the real scenario and limit the capability to capture complete illumination when encountered with complex environmental light, which is harmful to recover detailed face albedo. This observation motivates us to also refine the reflectance models in our method. Therefore, we propose to retain the Lambertian assumption and attribute all the complex reflectance into our detailed illumination map which is an extension of 3-band spherical harmonics.


    \section{Overview}
\label{sec:overview}

Given a single facial image as input, our goal is to reconstruct a 3D human face, with the emphasis on generating a high-fidelity and re-renderable facial texture that is complete, detailed and disentangled with illumination. 

To this end, we propose to first generate a prior albedo by a 
prior albedo generation module 
and enhance it with the facial texture unwrapped from the image by a detail refinement module, see Fig.~\ref{fig:overview}. 
A high-quality albedo dataset is needed to train the parametric model adopted in prior albedo generation module. Although Gecer et al.\cite{gecer2019ganfit} collect a dataset of facial textures, the samples in it are not re-renderable (thus can not be referred as albedo) because of the baked-in illumination.~\cite{lattas2020avatarme, lattas2021avatarme++} provide a facial reflectance dataset (including both diffuse and specular components) composed of 200 individuals with 7 different expressions. Though disentangled with illumination, their limited captured data makes it hard to keep performer idiosyncrasy when being used on face reconstruction task.
 To the best of our knowledge, no existing large-scale open-source facial albedo dataset is available to date. Therefore, we choose a traditional linear 3DMM\cite{paysan2009a} as our parametric model because a 3DMM albedo excludes the most of the illumination out. The other parametric models like\cite{gecer2019ganfit} can also be directly applied in our pipeline.

In our framework, the input facial image is first fed into the 3DMM encoder to produce 3DMM parameters (including identity, expression, and albedo), pose parameters and illumination parameters. Then, these regressed 3DMM parameters are passed to a fixed 3DMM decoder to acquire the prior albedo, 3DMM shape, camera pose matrix and coarse illumination. Next, we obtain the facial texture by unwrapping the input image according to the projected 3DMM shape. The unwrapped facial texture map and prior albedo map are fed into the detail refinement module, which is composed of a modified version of image-to-image translation network, to generate a detailed albedo map and a detailed illumination map. Finally, 
the detailed albedo and the detailed illumination combined with the 3DMM shape projected in camera space are rendered to the image space by a differentiable renderer. 

    \section{Methodology}
\label{sec:methodology}

\subsection{Prior Albedo Generation Module}
Our prior albedo generation module takes a 3DMM as proxy model and uses a convolutional neural network to estimate the parameters of facial geometry, albedo, pose and illumination. We adopt the state-of-the-art 3DMM coefficient regressor \cite{deng2019accurate} for the purpose. After the input image in fed to the network, a 257--dimensional parameter vector is regressed of which 80 is for shape, 64 for expression, 80 for albedo, 27 for illumination and 6 for camera pose. A textured 3D face model can be recovered by these parameters with a 3DMM decoder. As demonstrated in \cite{Blanz2002A}, the color and position of vertices can be represented as follows:
\begin{equation}
A=\overline{A}+F_{A}\Theta_{A},
\label{3DMM_albedo}
\end{equation}
\begin{equation}
S=\overline{S}+F_{I}\Theta_{I}+F_{E}\Theta_{E},
\label{3DMM_geometry}
\end{equation}
where $A\in \mathbb{R}^{3 \times N}, S\in \mathbb{R}^{3 \times N}$ represent the albedo and shape of the recovered 3D face with $N$ vertices. $\overline{A}\in \mathbb{R}^{3 \times N}$ and $\overline{S}\in \mathbb{R}^{3 \times N}$ symbolize the mean albedo and mean shape of the 3DMM. $F_{E}, F_{I}$ and $F_{A}$ are the fixed basis of expression, identity and appearance in the 3DMM model, while $\Theta_{E},\Theta_{I}$ and $\Theta_{A}$ represent the expression, identity and albedo coefficients, respectively. These variables are incorporated into the decoder by a direct matrix multiplication. 

In illumination reconstruction, we assume a Lambertian surface for the face material and utilize 3-band RGB spherical harmonics to represent the environmental illumination. The reflection of a vertex $s_i$ with vertex normal $n_i$ and albedo $a_i$ can be computed as follows:
\begin{equation}
T_{s_i}(n_i,a_i,\Theta_{L})=a_i\times \sum_{b=1}^{27}\Theta_{L}^{b}\Phi_b(n_i),
\label{SH}
\end{equation}
where $\Theta_{L}$ represents the illumination parameters and $\Phi_b\in \mathbb{R}^3\rightarrow \mathbb{R}$ represents the spherical harmonics function.

On the basis of the above reconstruction, we can derive the prior albedo by 3DMM albedo decoding and generate an image texture by sampling the corresponding projection pixels with 3DMM shape and pose parameters. After these two textures are obtained, they are projected onto 2D uv-space to accommodate with 2D CNN structure of detail refinement module. However, two problems are introduced in the image texture sampling procedure. First, the non-frontal face and occlusion problem in the facial images may cause incompleteness in the unwrapped texture.
Second, the inaccurate regression of pose and shape  parameters in many cases may also cause the image texture to generate artifacts, especially in the edge parts. Our detail refinement module introduced in the following section can resolve these two problems.

\subsection{Detail Refinement Module}
Our detail refinement module adopts an image-to-image translation network in the uv-space where the input includes two maps: a prior albedo map and a partial facial image texture. We first pad the unseen parts with Gaussian noises as being carried out in \cite{deng2018uv-gan} because filling the 'holes' in the unwrapped image texture is one of the goals of our detail refinement module. Then, these two maps are concatenated and fed into the refinement network which also produces two outputs: the detailed albedo map and the detailed illumination map. The detailed albedo map includes information about basic details in human faces, such as facial wrinkles, pores and etc. Meanwhile the detailed illumination map attempts to model spatially-complex environmental illumination that can not be captured in a previous coarse reconstruction. 
In the following, we elaborate on each part of the module and explain the specifically-devised loss functions. 

\subsection{Illumination Disentanglement}

\noindent\textbf{Illumination regularization. }
With respect to the illumination modeling, directly utilizing the coarse illumination in the prior generation module does not fully capture the complex illumination for the in-the-wild images. This situation will lead to the leakage of light information to the albedo which makes it not rerenderable. Therefore, a detailed representation for illumination is needed.

Given that our framework is trained in a self-supervised way, disentangling the illumination with albedo is not trivial. We take advantage of the coarse illumination spherical harmonics generated by the prior module and develop our illumination representation from it. We introduce a novel representation in the uv-space called spherical harmonics map which models a spherical harmonics illumination for every vertex in the face model. With the illumination map, we could not only model complex illumination in facial images, but also disentangle light by minimizing the distance with coarse spherical harmonics, which is named by illumination regularization loss. In particular, we represent the detailed illumination map $L_{detail}\in \mathbb{R}^{(B,27,H,W)}$ in the uv-space and directly regress it from the detail refinement module. Then, we regularize local illumination by devising a mean square error (MSE) loss to penalize detail and coarse illumination differences. The MSE loss is expressed as follows:
\begin{equation}
L_{reg-illu} = ||M_{uv} * (L_{detail} - L_{coarse})||^2,
\label{Reg Illu Loss}
\end{equation}
where $L_{coarse}\in \mathbb{R}^{(B,27,H,W)}$ is the coarse SH-illumination vector expanding to uv map size, and $M_{uv}\in \mathbb{R}^{(B,1,H,W)}$ stands for the uv-mask which represents the projected region of 3D face model in the uv-space. 

\noindent\textbf{Cross perceptual loss. }
\begin{figure}[tbp]
	\centering
	\includegraphics[width=1.0\linewidth]{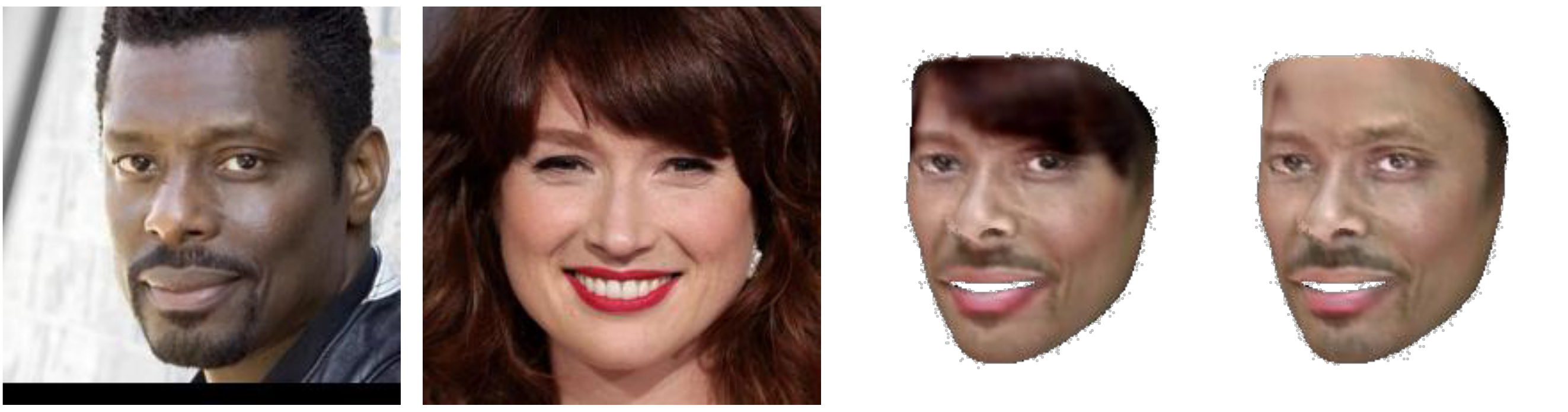}
	\caption{\textbf{Motivation of the cross perceptual loss.} From left to right are image of identity A, image of identity B, the rendered image constructed by detailed albedo of A in combination with detailed illumination of B without cross perceptual loss, the rendered image constructed by detailed albedo of A in combination with detailed illumination of B with cross perceptual loss}
	\label{fig:mid_results}
\end{figure}
Despite utilizing the illumination regularization loss mentioned above, we find in our experiments that a small amount of facial internal characteristics (such as wrinkles and beard) are mistakenly included in the illumination map, which means that albedo and illumination are not completely disentangled, leading to the loss of details in our detailed albedo map. Given that the attributes, such as wrinkles and beard are individual specific, we propose to utilize a cross-identity perceptual loss to conduct the further disentanglement. The motivation is stated below. We assume that the re-rendered image that combines person A's detailed illumination map and person B's detailed albedo map should have the same identity with person B in that the correctly-disentangled detailed illumination map would only include environmental illumination information in it. However, if the detailed illumination map is not entirely disentangled, which refers to including the facial attribute specific to person A, then it may change the identity of the rendered image. These two samples can be seen from Fig. \ref{fig:mid_results}. Owing to this observation, we utilize an illumination-irrelevant facial recognition network\cite{deng2018arcface} to distinguish whether the two images are of the same identity. The cross perceptual loss is represented as follows:
\begin{equation}
L_{cross-percp} = 1 - <ArcFace(I_r), ArcFace(I_{gt})>,
\label{Cross Percp Loss}
\end{equation}
where $ArcFace$ stands for the perceptual net, $I_r$ represents the rendered image with detailed illumination map A and detailed albedo map B, and $I_{gt}$ means the ground-truth facial image of B. We adopt the cosine distance as the measurement of the similarity between two normalized facial feature vectors.

We sum these two losses together with corresponding weights as our final illumination disentanglement loss:
\begin{equation}
L_{id} = \lambda_{id1}* L_{reg-illu} + \lambda_{id2}* L_{cross-percp}.
\label{Illu Disentangle Loss}
\end{equation}

\subsection{Albedo Regularization}
We utilize prior albedo and the intrinsic characteristics of albedo map to construct several regularization losses for obtaining a complete and re-renderable albedo from the input image. Here, we propose three losses : the symmetry loss, albedo smooth loss, and conditional GAN losses. 

\emph{(1) Symmetry loss:}
We propose a symmetry loss on the detailed albedo map. Given that human facial albedos are mostly symmetrical (especially when decoupled with light), we use this loss to regularize an unseen texture problem induced by the non-frontal face in the input image. Another advantage of the albedo symmetry loss is that it ensures robust albedo reconstruction in uneven scene illumination. The symmetry loss is expressed as follows:
\begin{equation}
L_{symm} = ||M_{uv} * (A_{detail} - \hat{A_{detail}})||^2,
\label{Symmetry Loss}
\end{equation}
where $\hat{A_{detail}}$ is the detailed albedo map flipped along the y-axis.

\emph{(2) Albedo smooth loss:}
We propose a smooth loss to regularize the detailed albedo map. We expect the detailed albedo map inherit this feature because the generated prior albedo is decoupled with illumination. We utilize local weighted smooth loss on the detailed albedo map to achieve this goal. In computing the local weights, we assume the detailed albedo map shares the same smoothness with the prior albedo map. Therefore, we use the local difference between pixels in the prior albedo map to compute the smoothness weights of the detailed albedo map. The albedo smooth loss is defined as: 
\begin{align}
L_{smooth} &= \sum_{i}\sum_{j\in N(i)} \omega_{i,j}||A_{detail}(i)-A_{detail}(j)||^2,\\
\omega_{i,j} &= exp(-\alpha * ||A_{prior}(i)-A_{prior}(j)||^2),
\end{align}
where $A_{detail}$ and $A_{prior}$ represent the detailed albedo map and the prior albedo map respectively, $N(i)$ indicates the neighbors of texel (pixel in uv-space) $p(i)$, $\omega_{i,j}$ represents the similarity of two texels, which is measured by a decreasing function of corresponding texels' difference in the prior albedo map. $\alpha$ is a super-parameter which we here choose 80 empirically\cite{tewari2019fml}. In the above equation, our albedo smooth loss penalizes more to those texels whose neighborhood difference shares less similarity between the detailed albedo map and the prior albedo map.

\emph{(3) GAN loss:}
Our devised GAN loss includes an $L_1$ distance loss and an adversarial loss \cite{pix2pix2017} to force the detailed albedo to share the same distribution as prior albedo map. The $L_1$ distance loss can be written as:
\begin{equation}
L_{L1} = |M_{uv} * (A_{detail} - {A_{prior}})|.
\label{L1 Loss}
\end{equation}
We then define the adversarial loss as:
\begin{align}
\label{Albedo Adversarial Loss Discriminator}
L_{GAN_D} = &E_{G(z)\in A_{detail}}log(1-D(G(z))) +\\
&E_{x\in A_{prior}}logD(x), \nonumber \\
\label{Albedo Adversarial Loss Generator}
L_{GAN_G} = &E_{G(z)\in A_{detail}}log(D(G(z))).
\end{align}
where $D$ symbolizes the discriminator to judge whether the generated albedo map falls on the support set of the prior albedo map distribution. $G(z)$ represents the detailed albedo map generator which means the whole framework. 

Our albedo regularization loss is then computed by combining above four loss terms with proper weights:
\begin{equation}
L_{ar} = \lambda_{ar1} * L_{symm} + \lambda_{ar2} * L_{smooth} + \lambda_{ar3} * L_{L1} + \lambda_{ar4} * L_{GAN_G}
\label{ar Loss}
\end{equation}

\subsection{Detail Preservation}

Besides above regularization losses, we also utilize basic reconstruction losses to facilitate high-fidelity reconstruction. These losses are all applied on the image space; thus a face mask is required for concentrating penalization of the differences on face regions in the images. We adopt the face parsing approach \cite{yu2018bisenet} to generate face masks before training. Coarse reconstruction may not be well-suited to image mask because of the inaccurate estimation of 3DMM and camera pose. Hence, we generate our final face mask by multiplying a pre-generated mask with projected face mask. The final face mask can be computed as:
\begin{equation}
\label{Face Mask}
M_{face} = M_{parsing} * M_{proj}.
\end{equation}

After face masks are obtained, we propose two reconstruction losses applied in the mask regions, which contain image gradient loss and image loss.

\emph{(1) Image gradient loss:} 
We now propose an image gradient loss to encourage the similarity between the re-rendered facial image gradient and the ground-truth facial image gradient for reconstructing facial details as authentic as possible. This loss is designed according to the assumption that the detail information can be mostly captured by image gradient. We define such gradient loss function as:
\begin{equation}
\label{Image Grad Loss}
L_{grad} = \sum||M_{face} * (Grad(I_{r}) - Grad(I_{gt}))||^2,
\end{equation}
where $M_{face}$ is the pre-extracted face mask,  $I_{r}$ and $I_{gt}$ are the rendered image and the input facial image, respectively; and $Grad$ represents the gradient operator. Specifically, we first calculate two directional gradients along the x-axis and y-axis, then compare them with corresponding ground-truth gradient maps. Finally, we obtain the summation.

Given that gradients are mostly relevant to facial inherent details such as wrinkles and pores, we should model them in the detailed facial albedo map rather than in the detailed illumination map. To achieve this goal, we add another layer to render a facial image with coarsely-reconstructed illumination and the detailed albedo map to obtain albedo-detail-only rendered image. We then apply our gradient loss on this image instead of the image that is rendered by using the detailed albedo and the detailed illumination to facilitate the detailed albedo map containing more facial detailed information.

\emph{(2) Image loss:} 
We also adopt image loss to penalize pixel difference between the rendered image and the input facial image, which can be expressed as follows:
\begin{equation}
\label{Image Loss}
L_{img} = \sum ||M_{face} * (I_{r} - I_{gt})||^2.
\end{equation}

We combine these two losses together to obtain our detail preservation loss:
\begin{equation}
L_{dp} = \lambda_{dp1} * L_{grad} + \lambda_{dp2} * L_{img}.
\label{dp Loss}
\end{equation}

\subsection{Network Architecture and Training Details}
We train our neural network on a public dataset CelebA~\cite{liu2015faceattributes} which is a large-scale facial attribute dataset that has more than 200K facial images collected from the internet. We separate the dataset into disjoint training data ($85\%$) and testing data ($15\%$). 
We pre-process the images by first generating 68-landmarks~\cite{feng2018joint} before feeding them into our detail generation network. Then, we utilize the generated landmarks to crop and scale the images to keep the human faces staying in the center of the images and resize them to $224 \times 224$. 
After pre-processing, these images are fed into our pre-trained prior albedo generation module. In this work, a 3DMM parameter regressing network and a fixed 3DMM decoder~\cite{paysan2009a} are utilized to obtain the prior albedo and other attributes (geometry, camera and illumination).
The camera parameters and 3DMM shape are leveraged to unwrap the texture from the input image. Next, the unwrapped texture and prior albedo (in uv-space) are concatenated and fed into the detail reconstruction network to acquire detailed albedo maps and detailed illumination maps. 

We adopt the ResNet-50~\cite{he2016deep} as the backbone network of our prior albedo generation module and pre-train it on 300W-LP~\cite{zhu2016face} following the state-of-art 3DMM reconstruction work \cite{deng2019accurate}. 300W-LP is a dataset that contains 122,450 facial images with a variety of head poses generated from the original 300W dataset by face profiling techniques. Similar to the training process in~\cite{deng2019accurate}, we train the network self-supervisedly with pixel-level, landmark-level and perceptual level discrepancy in combination with the parameter regularization loss. 

In the detail reconstruction network, we adopt the basic pix2pix network~\cite{pix2pix2017} as our backbone and abandon the skip-connection because it may cause the output to inherit the noise from the un-wrapped image texture. We also extend pix2pix with two decoders, one for detailed albedo generation and the other for detailed illumination modeling. Tables~\ref{tab:Encoder} and~\ref{tab:Decoder} summarize the network architectures of our encoder and decoder, respectively. The light decoder and albedo decoder share the same structure but with different output layers, where the light decoder outputs a 27-channel map while the albedo decoder outputs a 3-channel one.
Finally, We use the face mesh renderer~\cite{shi2020neutral} for differentiable rendering. We combine the loss functions mentioned above to train our network, which is expressed as follows:
\begin{equation}
L_{total} = L_{id} + L_{ar} + L_{dp}.
\label{Loss for all}
\end{equation}
The coefficients in those loss functions are chosen as $\lambda_{id1}$, $\lambda_{id2}$, $\lambda_{ar1}$, $\lambda_{ar2}$, $\lambda_{ar3}$, $\lambda_{ar4}$, $\lambda_{dp1}$, $\lambda_{dp2}$ : 1.0, 0.5, 5.0, 5.0, 1, 1.0, 0.001, 1.0, 5.0, where $\lambda_{ar3}$ is set as 0.0 after first epoch training.


\begin{table}[!tp]
\caption{Detailed architecture of our encoder. $Conv(c_{in},c_{out},k,s,p)$ refers to a convolution with $c_{in}$ input channels, $c_{out}$ output channels, kernel size $k$, stride $s$ and padding $p$. InsNorm stands for instance normalization, and we use the LeakyReLU for activation.}
\centering
\begin{tabular}{|c|c|}
\hline
\textbf{Encoder} & \textbf{Output Shape}\\
\hline
Conv(6,64,3,1,1), InsNorm, LeakyReLU & 256\\
\hline
Conv(64,64,4,2,1), InsNorm, LeakyReLU & 128\\
\hline
Conv(64,128,4,2,1), InsNorm, LeakyReLU & 64\\
\hline
Conv(128,256,4,2,1), InsNorm, LeakyReLU & 32\\
\hline
Conv(256,512,4,2,1), InsNorm, LeakyReLU & 16\\
\hline
\end{tabular}
\label{tab:Encoder}
\end{table}

\begin{table}[!tp]
\caption{Architecture of our decoders, where we use Transconv for transpose convolution. In the last layer a Conv operator is used to alleviate checkerboard artifacts caused by transpose convolution. The letter 'n' stands for the output channel, where we take a three-channel for albedo decoder and 27-channel for illumination decoder. }
\centering
\renewcommand\arraystretch{1.112}
\begin{tabular}{|c|c|}
\hline
\textbf{Decoder} & \textbf{Output Shape}\\
\hline
TransConv(512,256,4,2,1), InsNorm, ReLU & 32\\
\hline
TransConv(256,128,4,2,1), InsNorm, ReLU & 64\\
\hline
TransConv(128,64,4,2,1), InsNorm, ReLU & 128\\
\hline
TransConv(64,64,4,2,1), InsNorm, ReLU & 256\\
\hline
Conv(64,n,3,1,1) & 256\\
\hline
\end{tabular}
\label{tab:Decoder}
\end{table}

We trained our entire network end-to-end for 10 epochs using the Adam optimizer. 
The initial learning rate was set to $10^{-4}$ and reduced with attenuation coefficient of 0.98 every 1 epochs until we reached $10^{-5}$ to avoid overfitting.
The batch size was 16 and momentum was 0.9. The training task was completed in 2 days on a workstation with one Nvidia RTX-2080 TI GPU. 
Once trained, our network can process approximately 30 images per second in the inference stage. 
    \section{Experimental Results}
\label{sec:results}

We start with several experiments through visually inspecting our results on two well-known datasets to demonstrate the effectiveness of our proposed method. 
Then, we evaluate our algorithm qualitatively and quantitatively by performing a complete comparison with current state-of-the-art approaches. 
We further conduct ablation studies to provide a comprehensive evaluation of the individual components of our neural network. 
To facilitate future research, our source code is publicly released at \emph{https://github.com/YMX2022/3DFaceTexture}.

\begin{figure}[!tb]
\centering
  \includegraphics[width=1.0\linewidth]{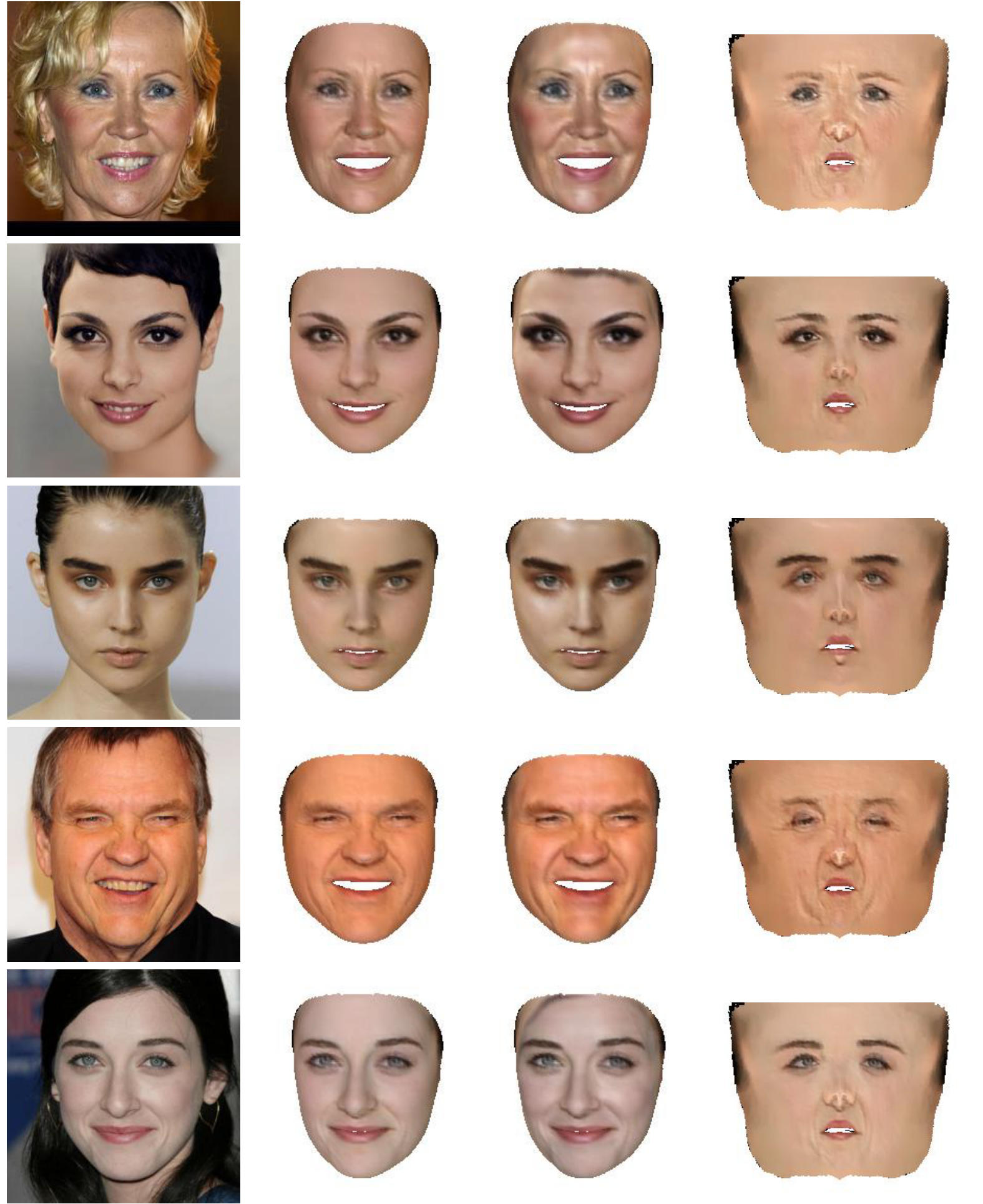}   
  \caption{\textbf{Our reconstruction results on CelebAHQ.} The first column is input, the second and third columns show the results generated by our detailed albedo combined with detailed illumination and coarse illumination respectively. The last column shows the detailed albedo in the uv-space.}
  \label{fig:qual_results}
\end{figure}

\begin{figure*}[!tb]
\centering
  \subfloat{\label{fig:a}\includegraphics[width=0.48\linewidth]{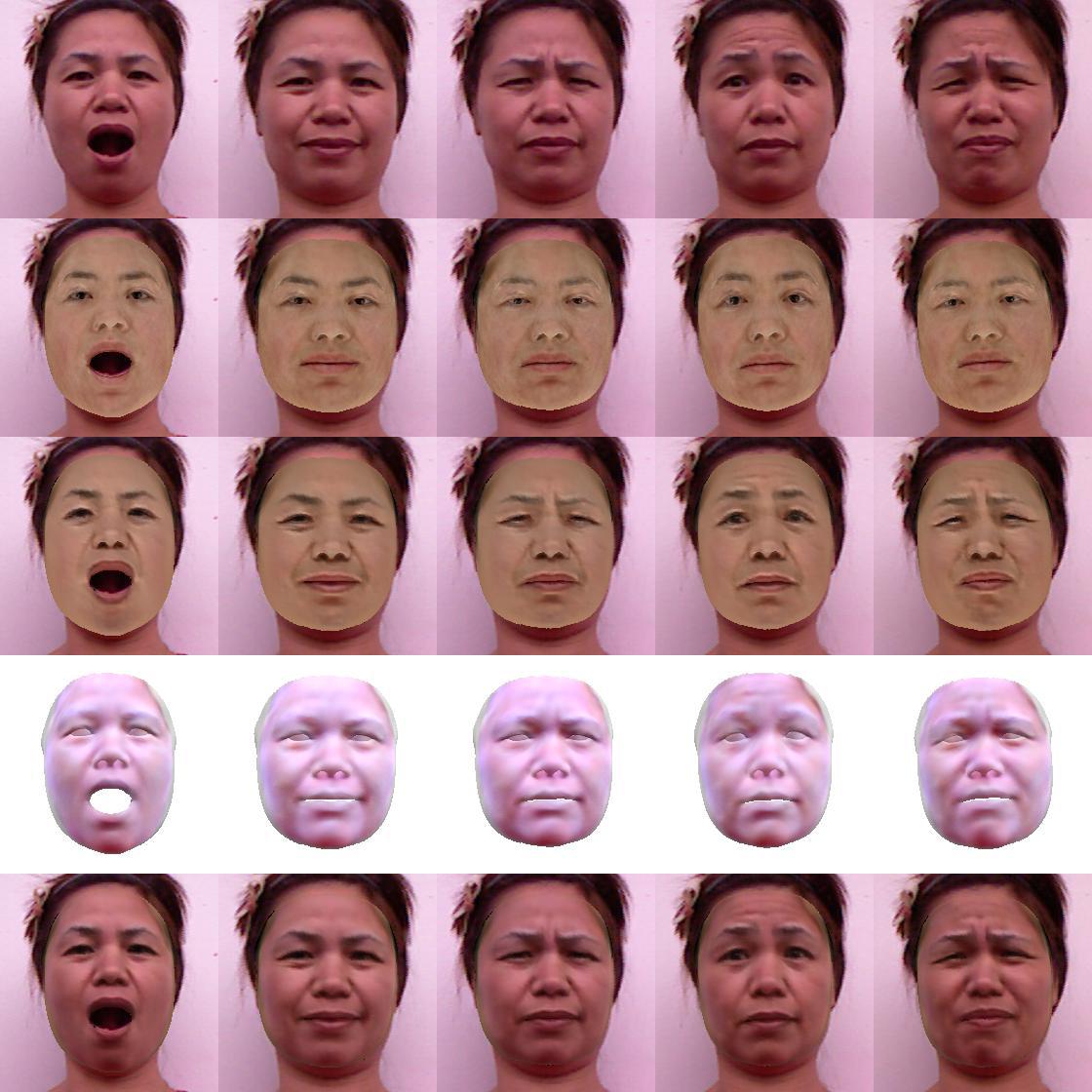}}\quad
  \subfloat{\label{fig:b}\includegraphics[width=0.48\linewidth]{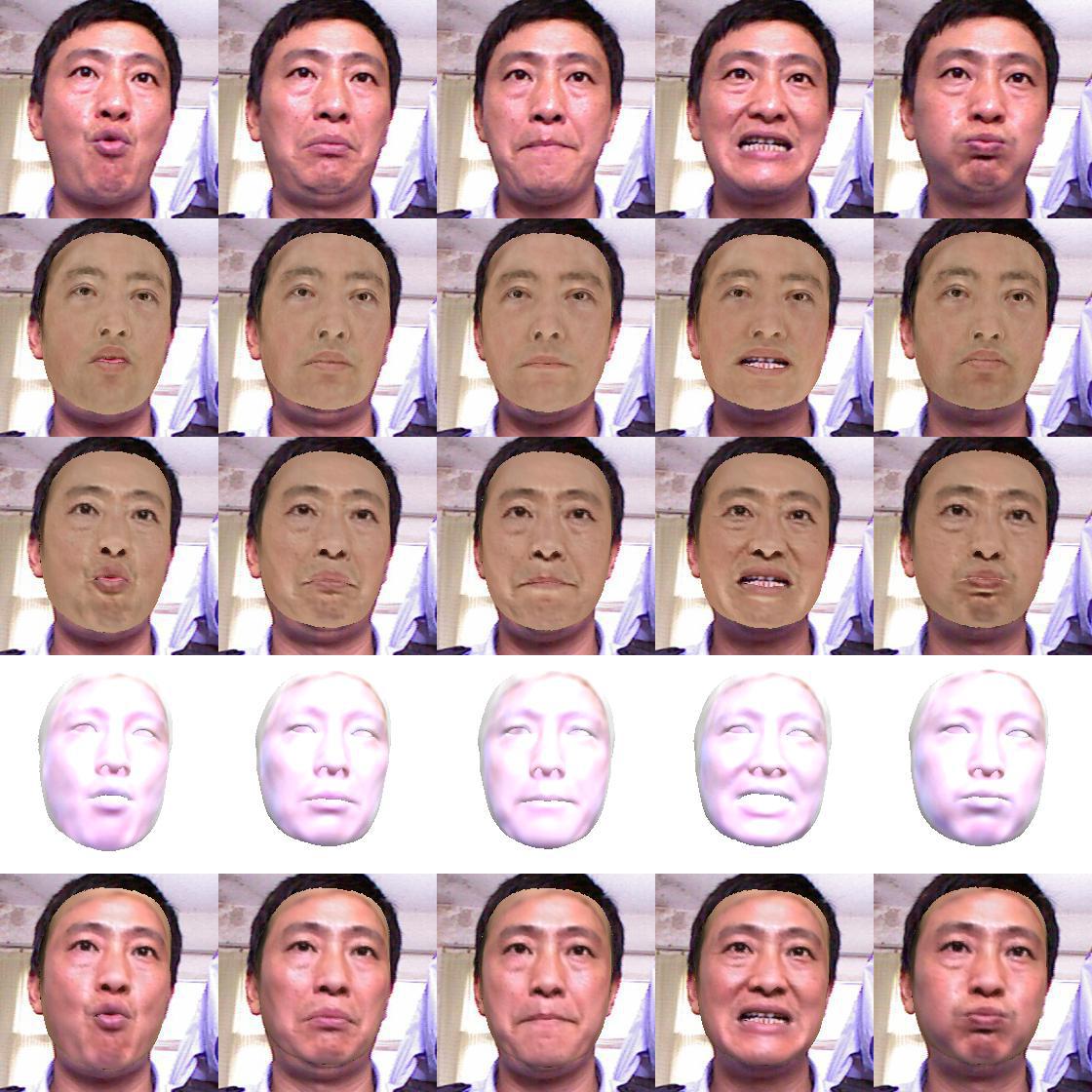}}
  \caption{\textbf{3D face reconstruction on two random FaceWareHouse (FWH)~\cite{cao2013facewarehouse} samples.} From top to bottom are input images, coarse reconstructed albedo overlay results, reconstructed detailed albedo overlay results, detailed illumination rendered to image space, detailed albedo combined with detailed illumination overlay results.}
  \label{fig:qual_results_1}
\end{figure*}

\subsection{Evaluation}

\noindent\textbf{Performance on CelebAHQ dataset.}
We first analyze the capability of our method by using the CelebAHQ database~\cite{CelebAMask-HQ}, which includes 30k $ 1024\times 1024$ facial images generated by applying super-resolution algorithm to a subset of CelebA images. Given that the images in CelebAHQ contain more details, 
we test our neural network on it to demonstrate whether our approach could capture details on high-definition images and achieve high-quality reconstructed albedos. Note that our neural network is only trained on the original CelebA.


Fig.~\ref{fig:qual_results} qualitatively shows our reconstruction results on several images randomly selected from CelebAHQ. 
Our proposed approach successfully keeps facial details in the reconstructed detailed albedo map, and there exists no reflection effects other than diffuse reflection in the albedo map or rendered results with coarse illumination. This phenomenon verifies that most of the environment illumination are explained by the detailed illumination, which leads to a clean diffuse facial albedo. In addition, our albedo regularization loss ensures that the detailed albedo map also exhibits smoothness and completeness which are beneficial to re-render applications. The detailed albedo generated by our network can be directly sent into a renderer with Lambertian reflector to achieve high-fidelity re-rendered results owing to these two characteristics.


\noindent\textbf{Performance on FaceWareHouse dataset.}
We use the FaceWareHouse (FWH)~\cite{cao2013facewarehouse} to test the network trained on CelebA dataset and demonstrate the generalization ability and robustness of our model. 
FWH includes 150 identities and each identity is captured with 20 specified expressions in a fixed environment. This dataset not only contains many challenging cases, such as high-frequency facial wrinkles generated by numerous expressions, but also includes the samples from the same identity in a fixed environment which leads to a rather stationary illumination condition. The second feature provides us a qualitative way to judge whether the environmental illumination is entirely explained by our proposed detailed illumination by inspecting if the detailed illumination maps of the same identity are close to each other. 

Fig.~\ref{fig:qual_results_1} shows the reconstruction results of two randomly selected identities. 
In the fifth row, the high-fidelity reconstruction results demonstrate that our model has good generalization ability and is robust enough to capture facial details in another dataset.
The second and third rows show the prior albedo and detailed albedo, respectively. The detailed albdeo shares strong appearance similarity to the prior coarse albedo while still preserving more details.
The similarity in the whole appearance proves that the prior albedo generated by 3DMM serves as an effective guidance in the detailed albedo reconstruction. 
The fourth row illustrates the detailed illumination that is rendered to the image space. The rendered detailed illumination images from the same identity have a similar appearance, which verifies the detailed illumination in our model correctly captures the environmental light.


\begin{figure}[!t]
	\centering
	\includegraphics[width=0.95\linewidth]{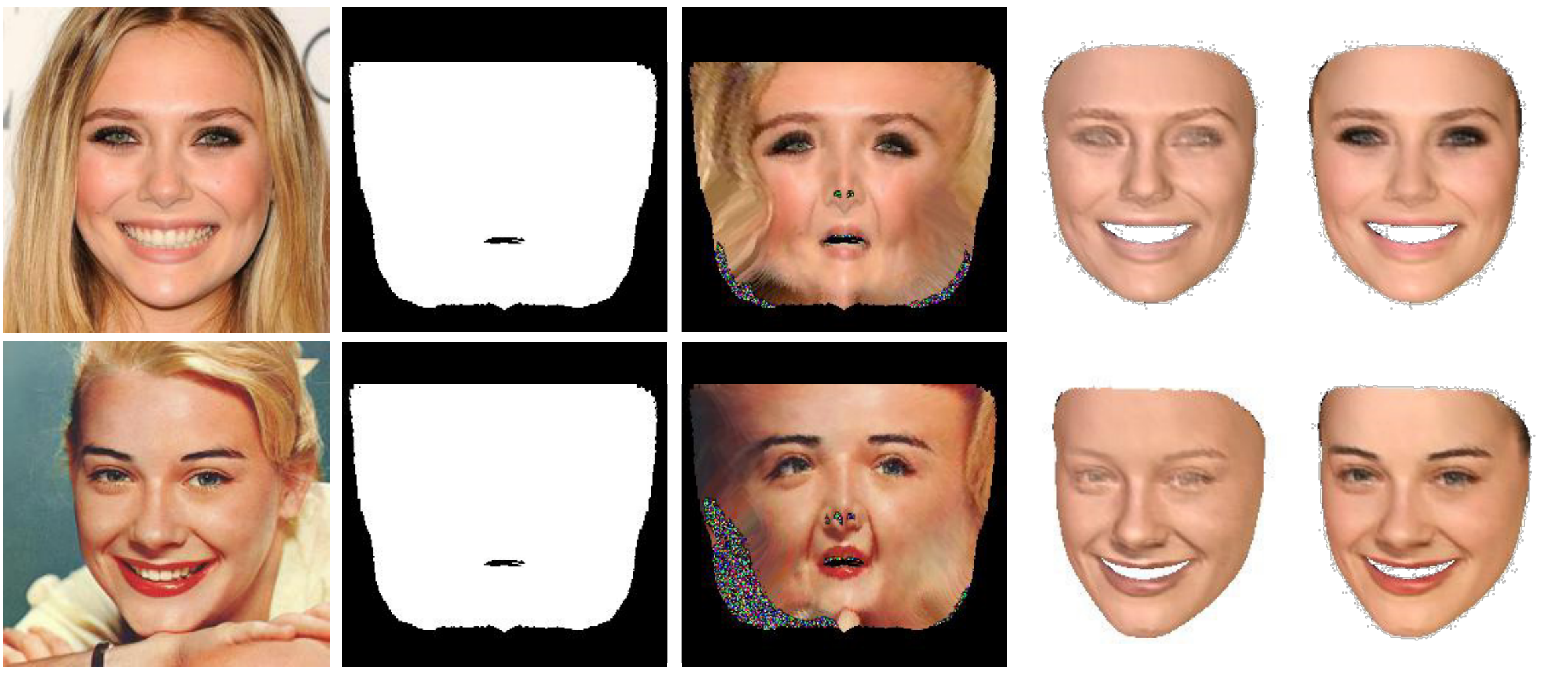}
	\caption{\textbf{Evaluation of prior albedo and detail refinement module.} The first three columns are input images, unwrapped image textures, and white facial albedo which replaces the coarse albedo. The right two columns show the output rendered images by using white albedo and original coarse albedo. The figure verifies the capability of our model to transfer as much details from the input texture. }
	\label{fig:preserve_cap}
\end{figure}

\begin{figure}[!tbp]
	\centering
	\includegraphics[width=1.0\linewidth]{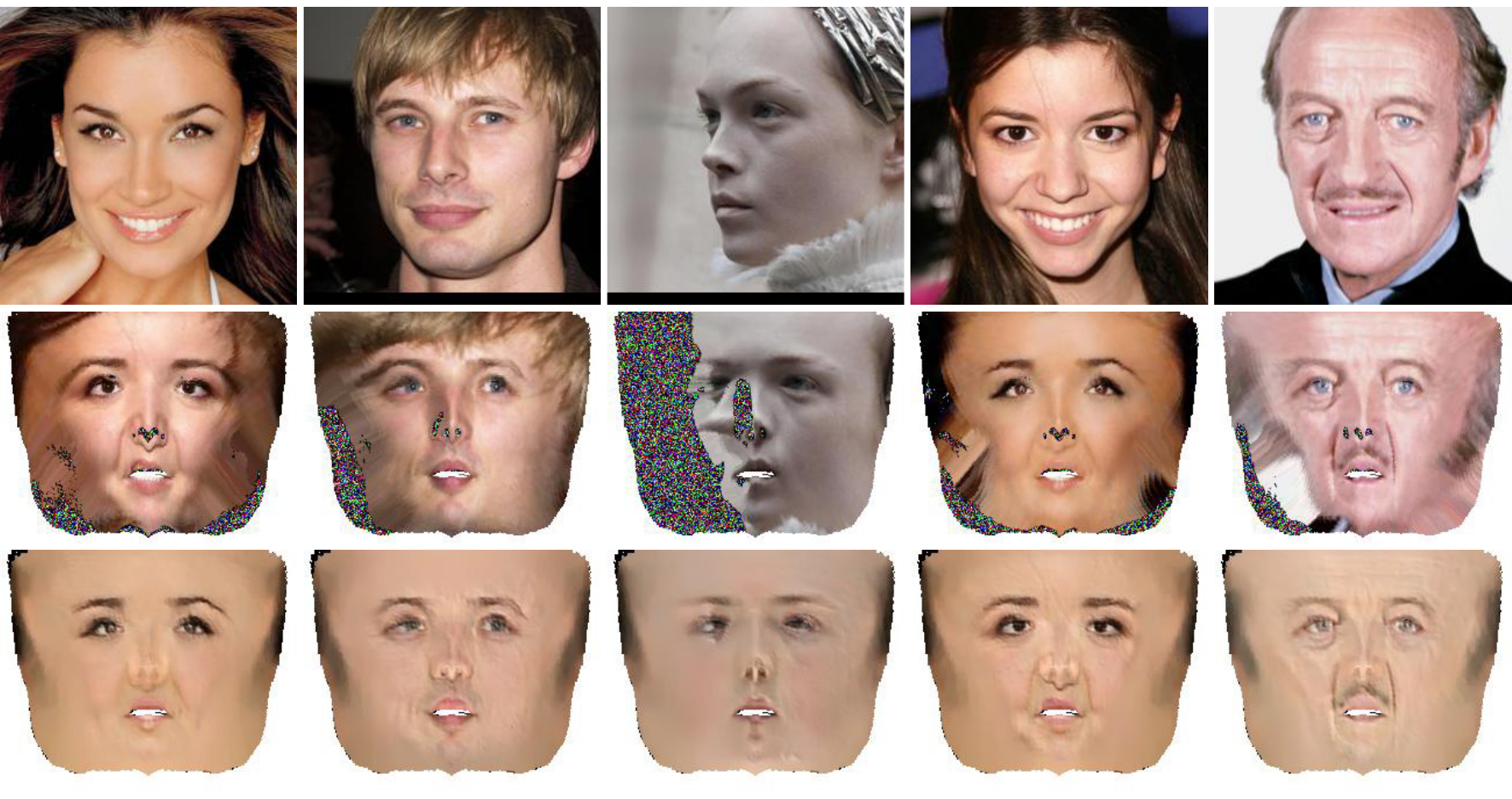}
	\caption{\textbf{Effectiveness of eliminating artifacts in unwrapped image texture.} From top to bottom are original input facial images, the unwrapped textures from input images and detailed albedos reconstructed by our method.}
	\label{fig:texture_syn}
\end{figure}

\begin{figure*}
\centering
\setlength{\fboxrule}{0.5pt}
\setlength{\fboxsep}{-0.01cm}
{     
    \begin{tabular}{cc}
    
    { \normalsize \raisebox{-.6\height}{\rotatebox{90}{Input image}}} & \raisebox{-.5\height}{\subfloat{
    \includegraphics[width=0.145\linewidth]{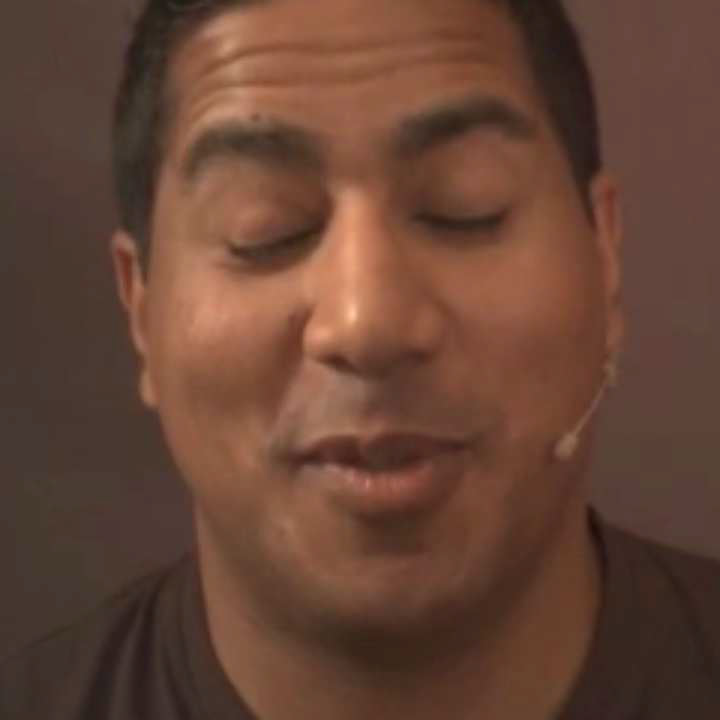}
    \includegraphics[width=0.145\linewidth]{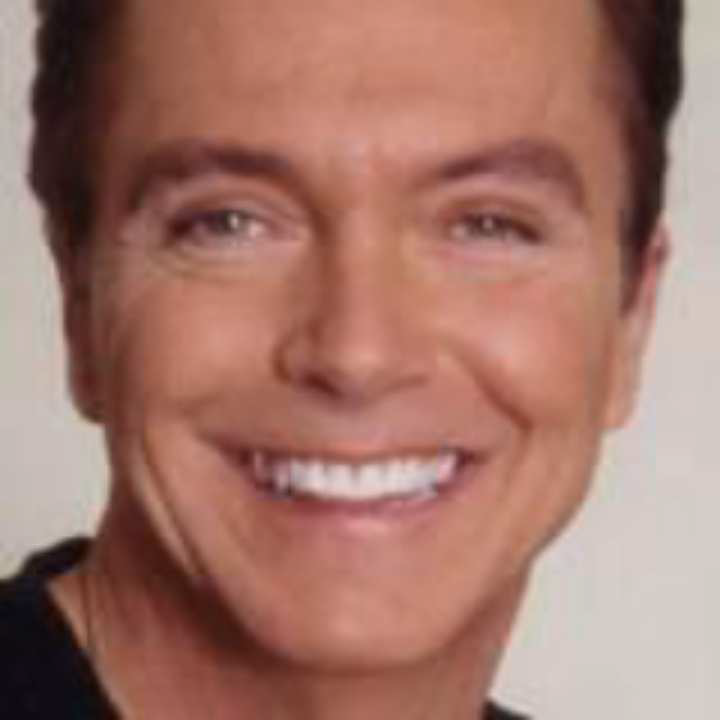}
    \includegraphics[width=0.145\linewidth]{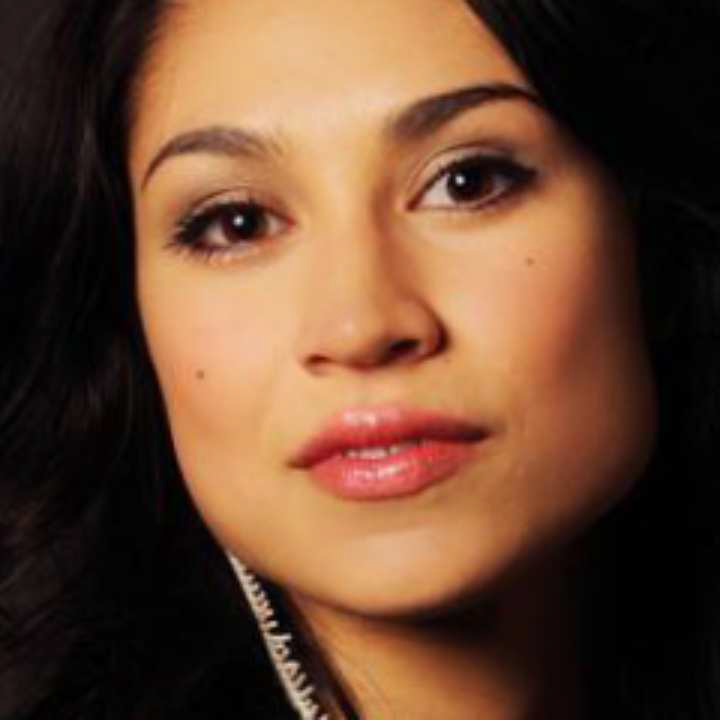}
    \includegraphics[width=0.145\linewidth]{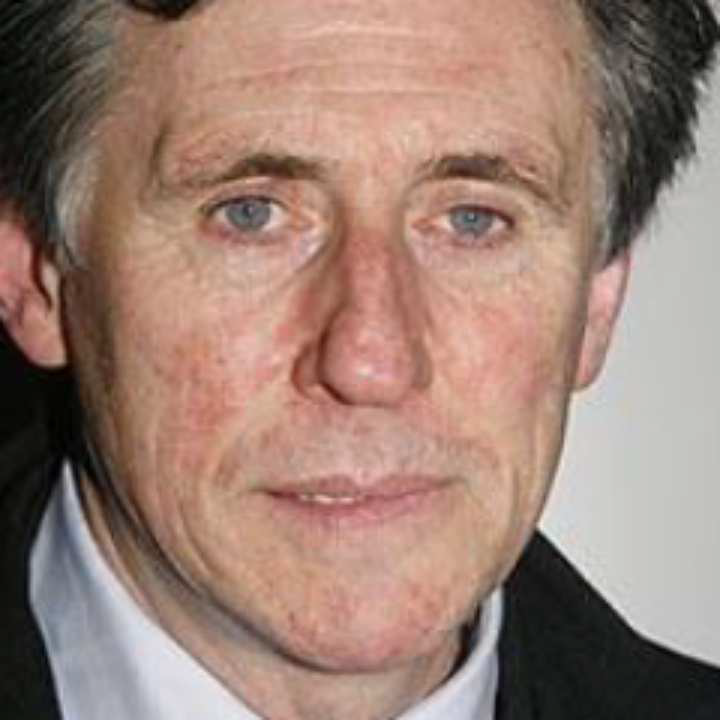}
    \includegraphics[width=0.145\linewidth]{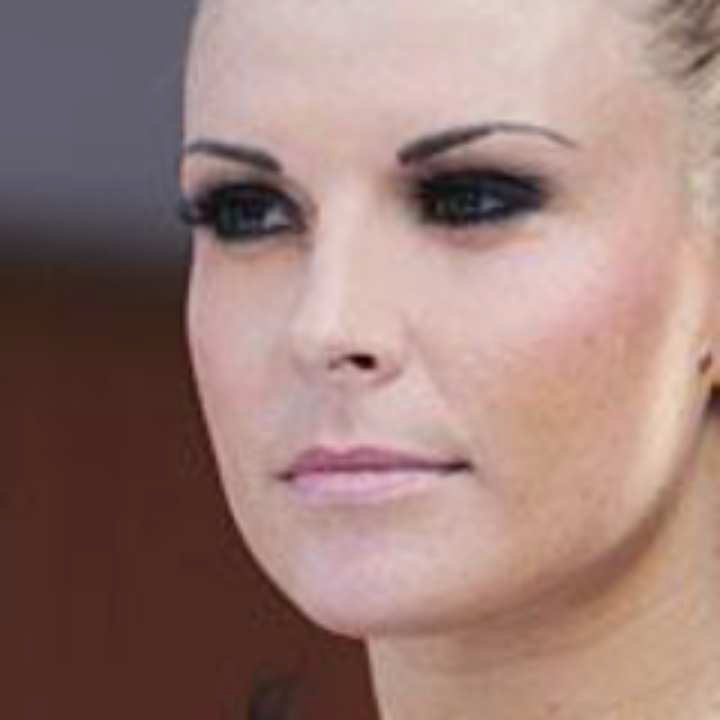}
    \includegraphics[width=0.145\linewidth]{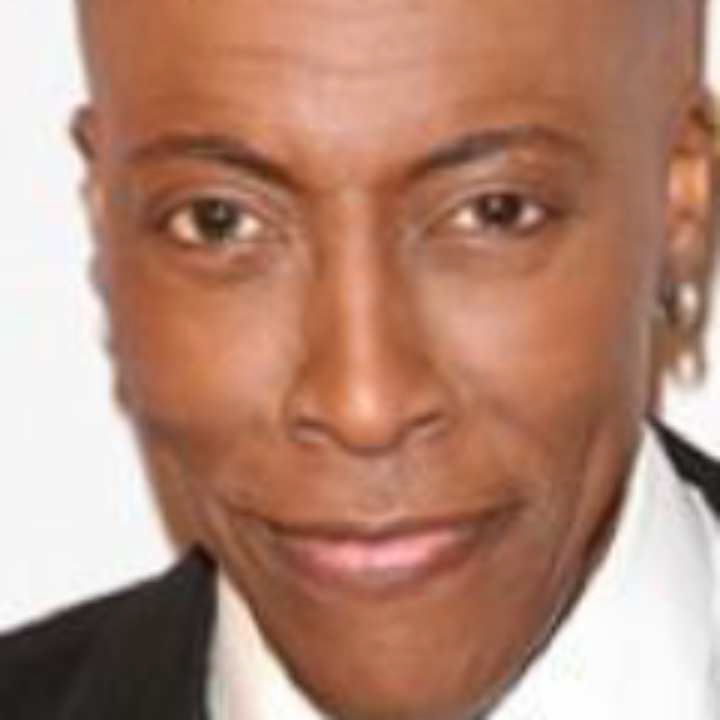}}}\\
    
    { \normalsize \raisebox{-.6\height}{\rotatebox{90}{Ours\_DI}}} & \raisebox{-.5\height}{\subfloat{
    \includegraphics[width=0.145\linewidth]{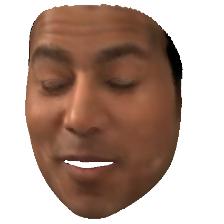}
    \includegraphics[width=0.145\linewidth]{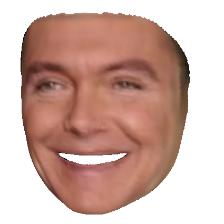}
    \includegraphics[width=0.145\linewidth]{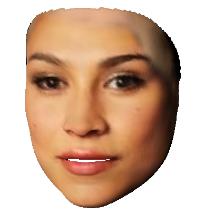}
    \includegraphics[width=0.145\linewidth]{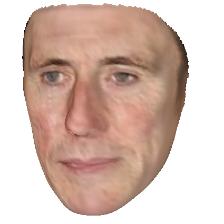}
    \includegraphics[width=0.145\linewidth]{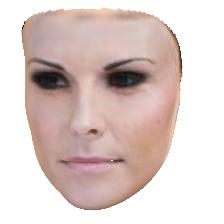}
    \includegraphics[width=0.145\linewidth]{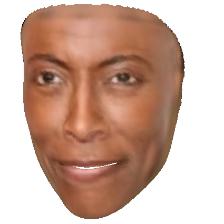}}}\\
    
    { \normalsize \raisebox{-.6\height}{\rotatebox{90}{Ours\_CI}}} & \raisebox{-.5\height}{\subfloat{
    \includegraphics[width=0.145\linewidth]{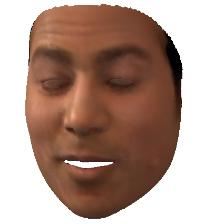}
    \includegraphics[width=0.145\linewidth]{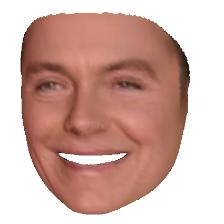}
    \includegraphics[width=0.145\linewidth]{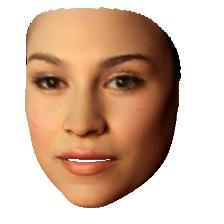}
    \includegraphics[width=0.145\linewidth]{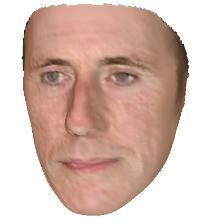}
    \includegraphics[width=0.145\linewidth]{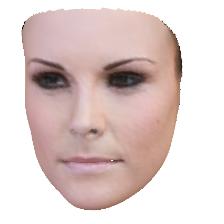}
    \includegraphics[width=0.145\linewidth]{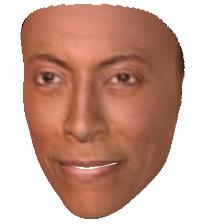}}}\\
    
    { \normalsize \raisebox{-.6\height}{\rotatebox{90}{Lin et al.\cite{jiangke2020towards}}}} & \raisebox{-.5\height}{\subfloat{
    \includegraphics[width=0.145\linewidth]{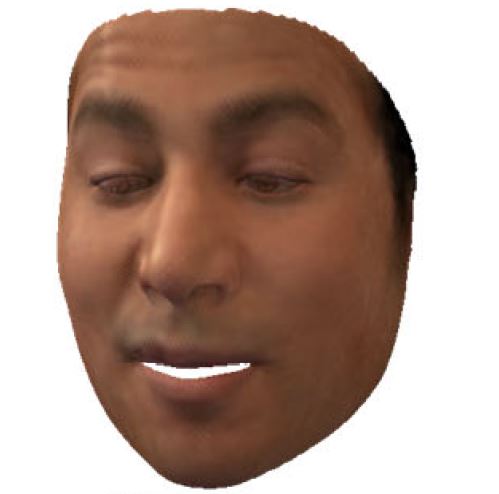}
    \includegraphics[width=0.145\linewidth]{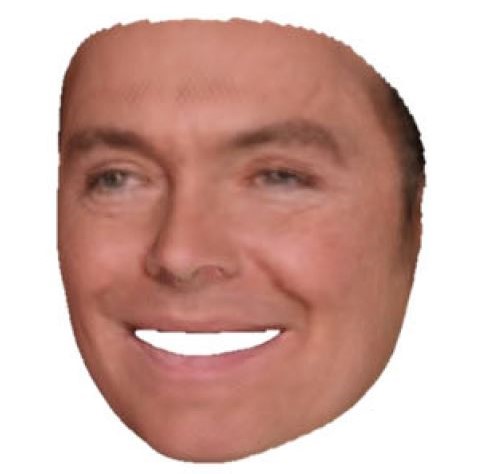}
    \includegraphics[width=0.145\linewidth]{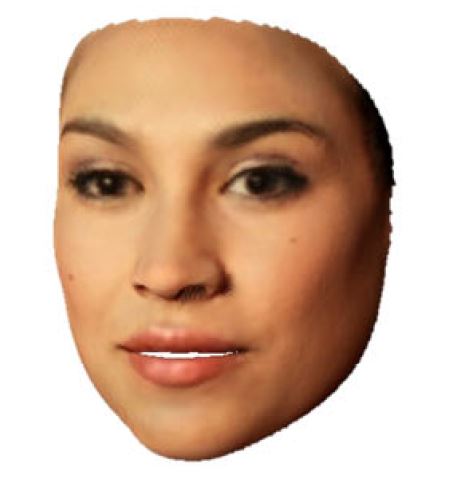}
    \includegraphics[width=0.145\linewidth]{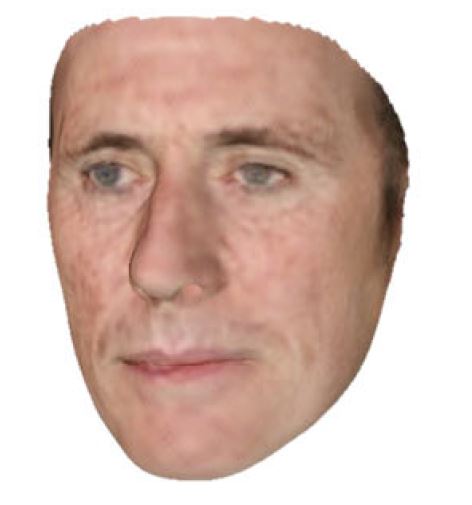}
    \includegraphics[width=0.145\linewidth]{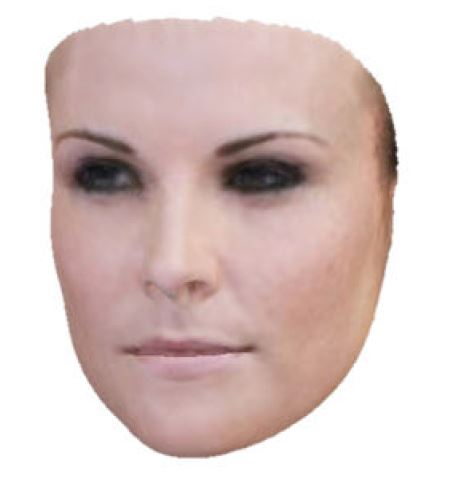}
    \includegraphics[width=0.145\linewidth]{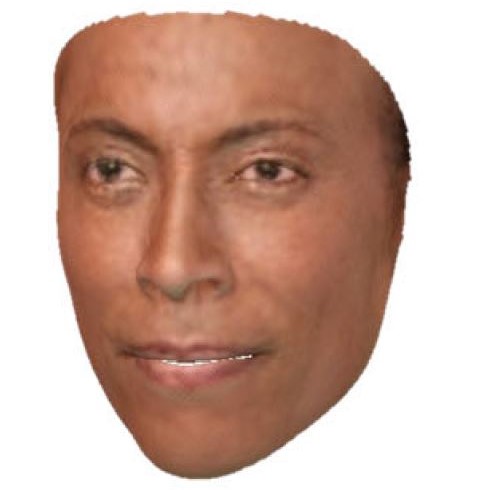}}}\\
    
    { \normalsize \raisebox{-.6\height}{\rotatebox{90}{Gecer et.al\cite{gecer2019ganfit}}}} & \raisebox{-.5\height}{\subfloat{
    \includegraphics[width=0.145\linewidth]{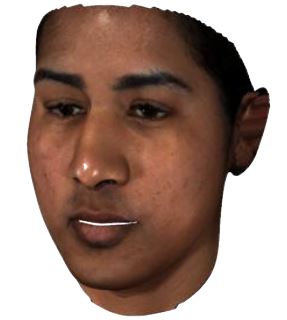}
    \includegraphics[width=0.145\linewidth]{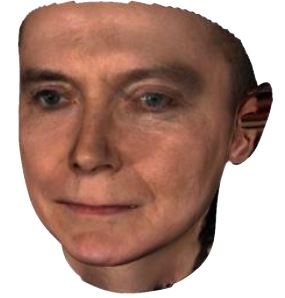}
    \includegraphics[width=0.145\linewidth]{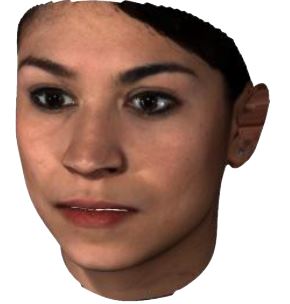}
    \includegraphics[width=0.145\linewidth]{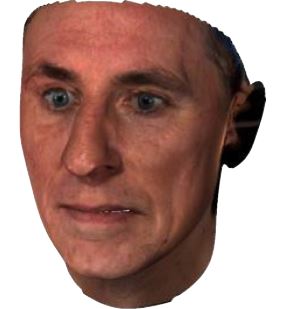}
    \includegraphics[width=0.145\linewidth]{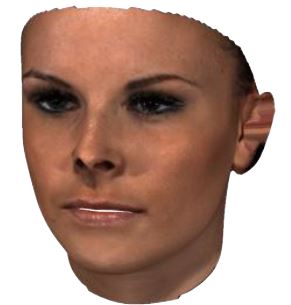}
    \includegraphics[width=0.145\linewidth]{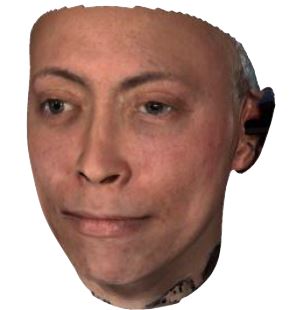}}}\\
    
    { \normalsize \raisebox{-.6\height}{\rotatebox{90}{Deng et.al\cite{deng2019accurate}}}} & \raisebox{-.5\height}{\subfloat{
    \includegraphics[width=0.145\linewidth]{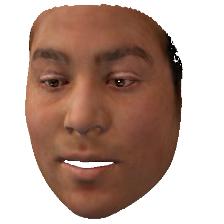}
    \includegraphics[width=0.145\linewidth]{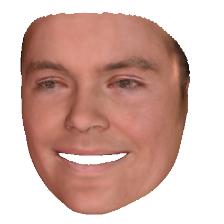}
    \includegraphics[width=0.145\linewidth]{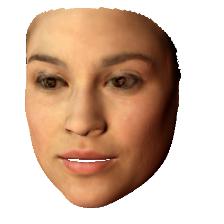}
    \includegraphics[width=0.145\linewidth]{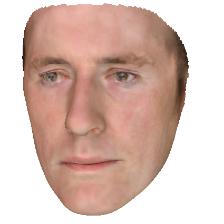}
    \includegraphics[width=0.145\linewidth]{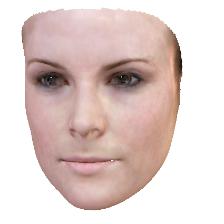}
    \includegraphics[width=0.145\linewidth]{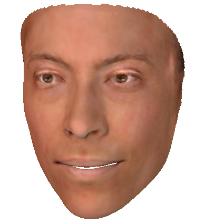}}}\\
    
    { \normalsize \raisebox{-.6\height}{\rotatebox{90}{Genova et.al\cite{genova2018unsupervised}}}} & \raisebox{-.5\height}{\subfloat{
    \includegraphics[width=0.145\linewidth]{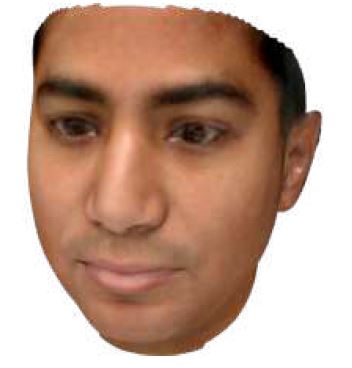}
    \includegraphics[width=0.145\linewidth]{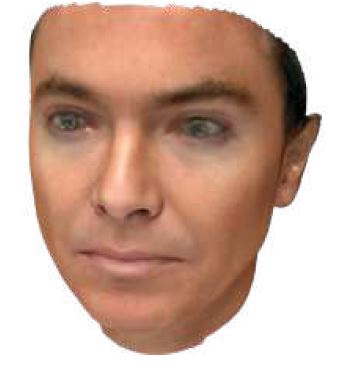}
    \includegraphics[width=0.145\linewidth]{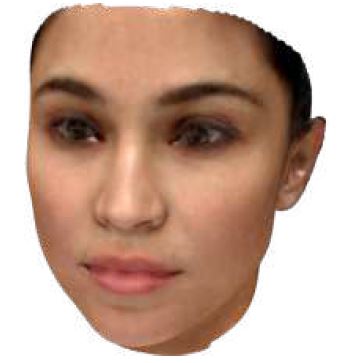}
    \includegraphics[width=0.145\linewidth]{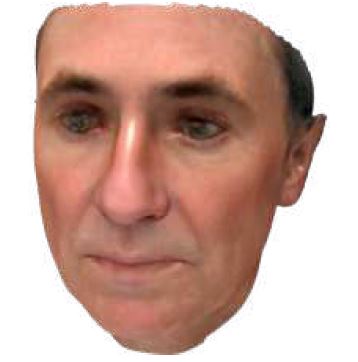}
    \includegraphics[width=0.145\linewidth]{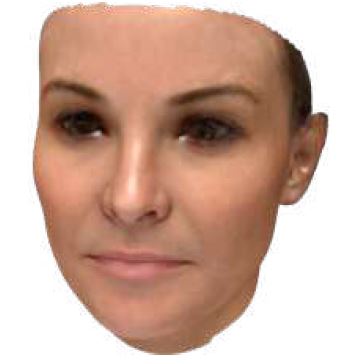}
    \includegraphics[width=0.145\linewidth]{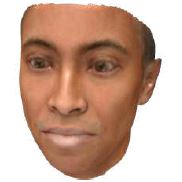}}}\\
    
    \end{tabular}
    \caption{\textbf{Qualitative comparison to other competitive methods.} The first row is the input images while the remaining rows show the reconstruction results of all methods. Our reconstructions are shown in the second and third rows where Ours\_DI stands for reconstruction with detailed illumination and Ours\_CI means reconstruction with coarse illumination.}
    \label{fig:compared_all}
}
\end{figure*}

\noindent\textbf{Evaluation of prior albedo and detail refinement module.}
Given that the detail refinement module in our framework takes two inputs, namely the prior albedo generated by 3DMM and the unwrapped texture from the input image, we are interested in exploring what these two inputs are responsible for in the detail refinement module. Accordingly we conduct an experiment by substituting the input prior 3DMM albedo by a white map (so it includes no prior information) to evaluate the effect of the inputs. Fig.~\ref{fig:preserve_cap} shows the experimental results, where the rendered image with white map absorbs most details in the input but loses the appearance consistency with the original image. This phenomenon indicates that our detail refinement module takes these two inputs independently, where the prior albedo guides entire appearance generation and the unwrapped texture is responsible for the detail supplement.


\noindent\textbf{Texture artifact removal.}
We now evaluate the capability of our detail refinement module in dealing with the textures containing artifacts. The two main artifacts that exist in facial images are show in the second row of Fig.~\ref{fig:texture_syn}. First, the non-frontal face images would lead to incompleteness in the unwrapped image texture. Second, the geometry parameters (including camera pose and shape parameters) regressed from the coarse reconstruction step are not accurate in many cases, which would result in severe stripe-like artifacts in the unwrapped texture, especially in the boundaries of human face. Our detail refinement module can remove these two kinds of artifacts and produce a smooth, complete and high-fidelity albedo (Fig.~\ref{fig:texture_syn}). This phenomenon is mainly due to the introduction of prior albedo and our designed albedo regularization loss, which endow the final reconstructed albedo with completeness and smoothness.

\subsection{Qualitative Comparison}
For qualitative comparison, we first compare our approach against recent learning-based texture reconstruction methods~\cite{jiangke2020towards,gecer2019ganfit,deng2019accurate,genova2018unsupervised}. 
Then, we focus on the qualitative evaluation in extreme illumination condition and compare our reconstructed albedo with advanced facial texture generation method~\cite{gecer2019ganfit}. Finally, we compare our albedo reconstruction performance with~\cite{yamaguchi2018high-fidelity} which shares similar goal with ours whereas utilizing a self-collected high-fidelity dataset.

\noindent\textbf{Comparison on MOFA data.}
Fig.~\ref{fig:compared_all} compares the reconstruction results on a subset of MOFA test dataset\cite{tewari2017mofa}.
Lin et al.~\cite{jiangke2020towards} aim to reconstruct high-fidelity facial texture from a single image self-supervisedly. They achieve this goal by also first utilizing the coarse albedo generated from 3DMM then fine-tuning it by a face feature extracted from the image according to a pre-trained face recognition network (FaceNet~\cite{schroff2015facenet}). They leverage a GCN as and the fine-tune module and represent the albedo on a mesh level (per-vertex albedo). Although looks similar, the target of our method is not exactly the same as theirs because we expect our reconstructed albedo to exhibit not only high-fidelity but also smooth, complete and decoupled from environmental illumination. Accordingly, our reconstructed albedo can be directly re-rendered by any renderer with a Lambertian reflector. Therefore, the third and fourth rows illustrate that,
our albedo renderers capture more internal facial details than \cite{jiangke2020towards} while leaving the complex environmental illumination and other reflection effects (such as specular reflection) aside from diffuse reflection to our detailed illumination map. Considering the main differences in implementation, our network has an additional illumination modeling module that accounts for the complex environmental illumination and we get rid of the pre-trained face feature extraction network used in~\cite{jiangke2020towards}. It is also worth noticing that the prior albedo generation module in our framework can be directly replaced by any generation model which leads to a strong extendibility of our work.

As a generation-based model, Gecer et al.~\cite{gecer2019ganfit} capture 10,000 high-resolution human facial textures in the uv-space and train a progressive growing GAN to model the distribution human face texture. They leverage this progressive GAN as the generative model and utilize fitting-based paradigm to estimate the parameters in latent space. The fifth row in Fig. \ref{fig:compared_all} show that the reconstruction results lose much fidelity from the original input face and are not alike in appearance. The phenomenon is especially severe for in-the-wild situations. This phenomenon could be attribute to the limited number of captured data that cannot span all of the face texture space. 
By constrat, our model utilizes a less expressive generative model (Basel Face Model 2009) but achieves better reconstruction fidelity.

Deng et al.~\cite{deng2019accurate} and Genova et al.~\cite{genova2018unsupervised} are the two representative 3DMM-based facial reconstruction methods that are also trained in the self-supervised way on in-the-wild facial image datasets. The sixth and seventh rows show the reconstruction results of these methods. The 3DMM-based reconstruction approach has the advantages of obtaining smoothness and completeness but loses most of the details in the face, which produces much less convincing results than our approach.

\begin{figure}[!tb]
\centering
  \includegraphics[width=1.0\linewidth]{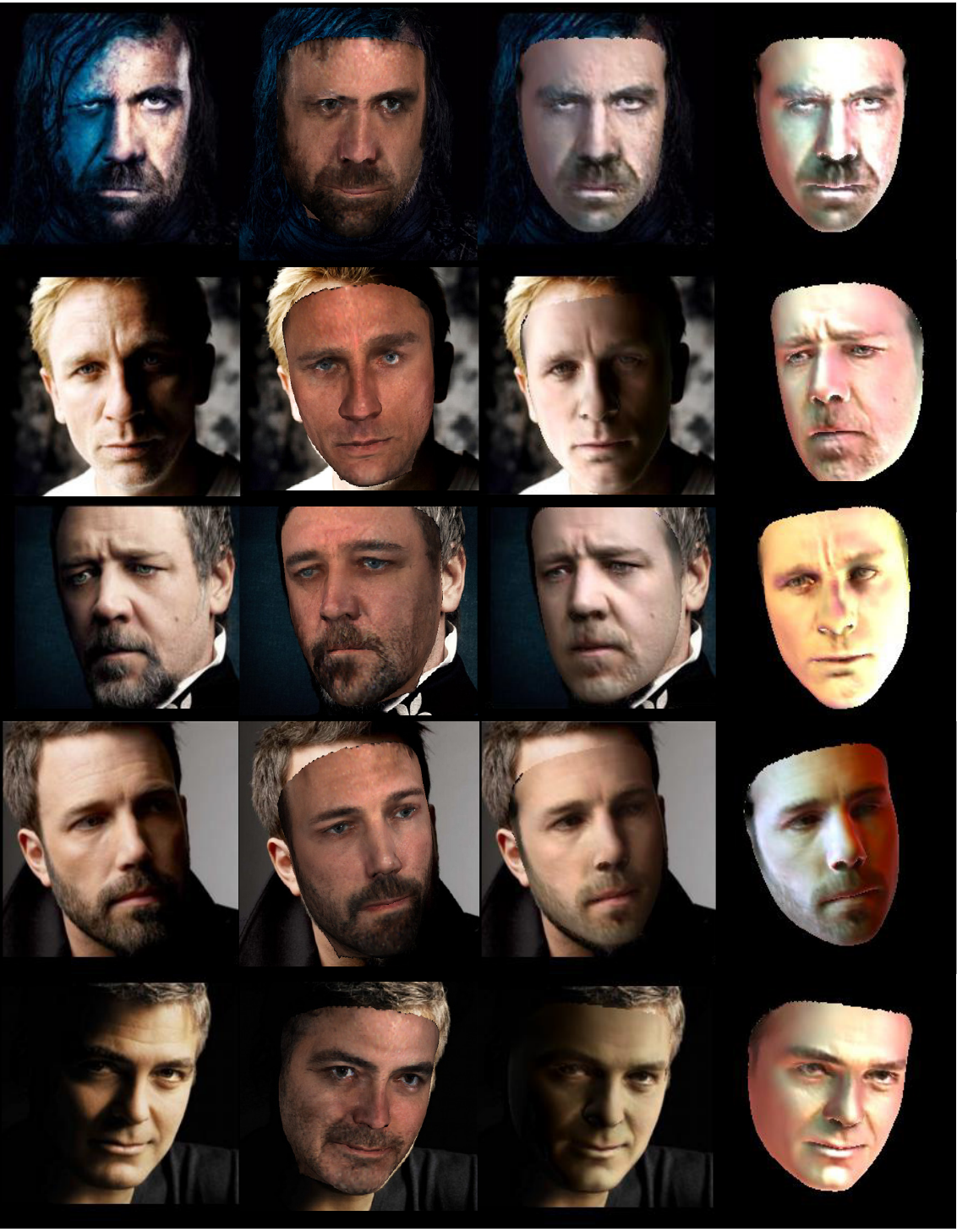}   
  \caption{\textbf{Comparison to Ganfit~\cite{gecer2019ganfit} in extreme lighting.} From left to right are input images, reconstructed albedo of~\cite{gecer2019ganfit}, our generated albedo, and our re-illuminated results using a different illumination condition. }
  \label{fig:compared_extreme}
\end{figure}

\noindent\textbf{Comparison on extreme illumination data.}
Facial images under extreme illumination condition, including uneven lighting or shadow, are commonly encountered in real-world applications. Due to the loss of information and low quality representation of illumination (three-band Spherical Harmonics), reconstructing high-fidelity 3D face under such circumstances is still challenging thus reconstruction-based methods always fail to reconstruct complete face albedos. 
Meanwhile, a generation-based approach can deal with extreme lighting because they map the facial texture space to a latent space with a support set.
Our proposed model merging these two methods together should also have the capability to reconstruct convincing face albedos from the facial images under extreme illumination.
In Fig. \ref{fig:compared_extreme}, we compare our reconstruction results in extreme lighting to Ganfit~\cite{gecer2019ganfit}.
The second and third columns show that our albedo decouples the complex environmental illumination and outperforms \cite{gecer2019ganfit} by preserving more facial details from the input face. This phenomenon is because of the cooperation between the prior albedo generation module with the detail refinement module of our framework. The former module is responsible for generating the guided albedo and the latter can complement details upon it. 
As a result, our method not only inherits the advantage of generation-based methods which maintain the diffuse texture smooth in the whole but also has the ability to preserve the details as in a reconstruction-based method. We outperform Ganfit and achieve more convincing results.
The fourth column shows re-illuminated results according to our reconstructed albedo, where we apply different illumination conditions to the reconstructed albedo and render it to images where the illumination is randomly selected from a face illumination prior database~\cite{schneider2017efficient}. The re-illuminated results are rather realistic and keeps the identity information of the original image.

\begin{figure}[!tb]
\centering
  \includegraphics[width=1.0\linewidth]{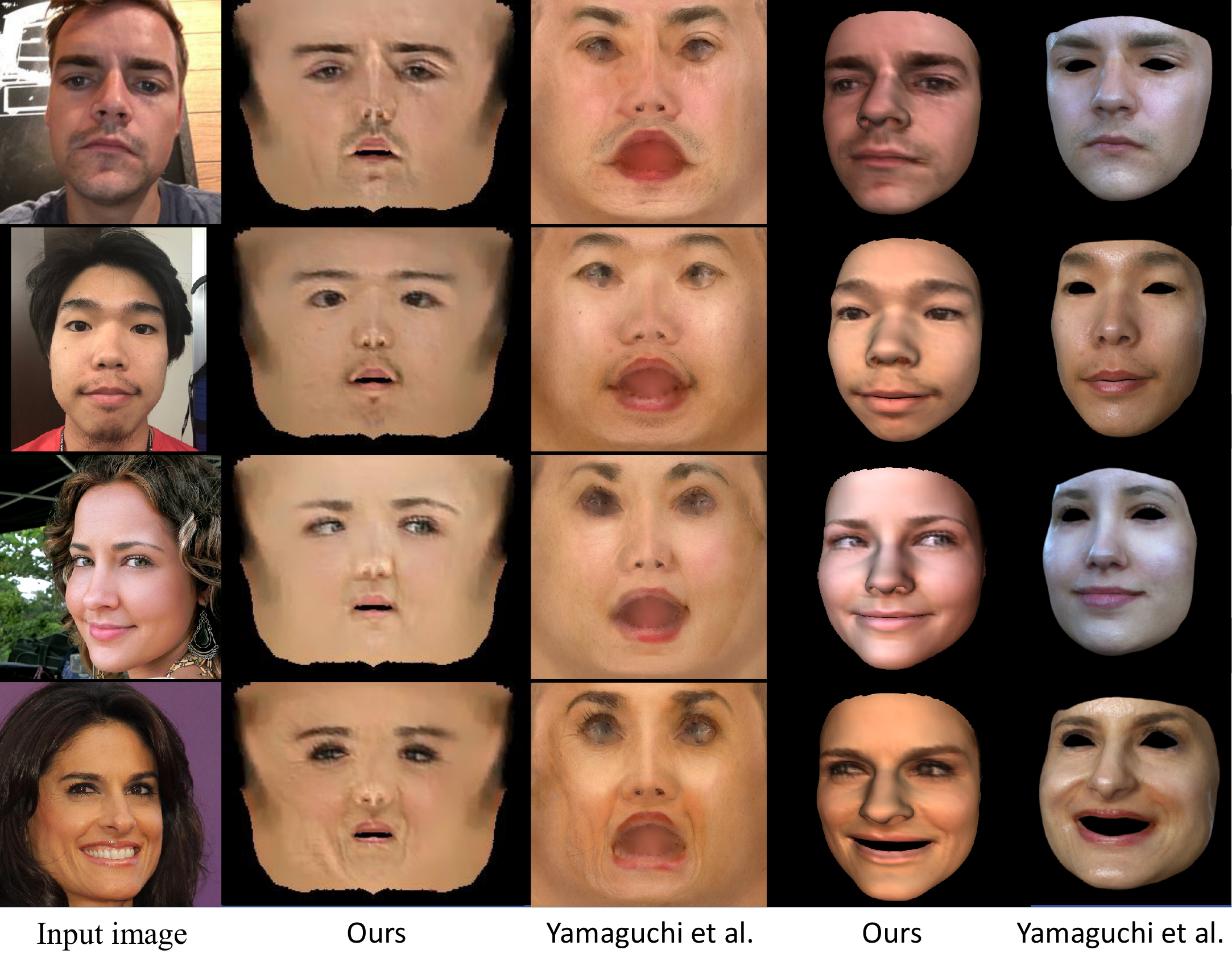}   
  \caption{\textbf{Comparison to Yamaguchi et al.~\cite{yamaguchi2018high-fidelity} for albedo reconstruction.} The first column shows input images; the second and third columns are the reconstructed albedos of our approach and~\cite{yamaguchi2018high-fidelity}; the last two columns display the rendered images with different viewpoints and illumination conditions.}
  \label{fig:compared_albedo}
\end{figure}

\noindent\textbf{Comparison on albedo reconstruction from a single image.}
To evaluate the quality of our reconstructed (diffuse) albedo, we compare with the state-of-the-art method~\cite{yamaguchi2018high-fidelity}. As shown in Fig.~\ref{fig:compared_albedo}, both~\cite{yamaguchi2018high-fidelity} and our approach can decouple environmental illumination well. However, thanks to the novel illumination representation and disentanglement loss, our reconstructed albedos keep more performers' idiosyncrasy than~\cite{yamaguchi2018high-fidelity}, which can be observed from the nasolabial folds from the third and fourth rows.

\subsection{Quantitative Comparison}
For quantitative comparison, we mainly focus on the criteria for measuring the image-level difference.
First, L1 distance loss is applied as the basic pixel-level criterion. Then, we utilize two commonly-used image similarity criteria, namely the structural similarity index measure (SSIM) and peak signal-to-noise ratio (PSNR), to evaluate the similarity between the rendered face image and original input face image. With regard to the human face problem, we also leverage two well-known pre-trained face recognition networks as maps from image space to feature space and evaluate the difference between rendered face image and input face image in the facial feature space. The two facial recognition networks we adopted are $\textit{LightCNN}$~\cite{wu2018light} and $\textit{evoLVe}$~\cite{zhao2019multi}, since their state-of-the-art performance and widely acceptance~\cite{jiangke2020towards}. In summary, we calculate the difference between two face images in both pixel-level (including L1 distance loss, PSNR and SSIM) and face feature-level (including LightCNN and evoLVe).

The numerical statistics for each method are reported in Table~\ref{tab:quant}, where the competing methods we choose are state-of-the-art ones trying to reconstruct details in albedo.
The table illustrates that our reconstruction results with detailed illumination are better than those of the competing algorithms. Besides, our framework also achieves competitive results by using only detailed albedo combined with coarse illumination, which further demonstrates that our detailed albedo is able to capture most facial details in the input image. 


\setlength{\tabcolsep}{5pt}
\begin{table}[!t]  
\renewcommand{\arraystretch}{1.1}
\centering
\caption{\textbf{Quantitative comparison on CelebA dataset.} Ours\_OD means the rendered image constructed by our detailed albedo combined with our detailed illumination, while Ours\_OC means the rendered image by using coarse illumination. 
The best result of each measurement is marked in \textbf{bold} font.
The symbol of '/' means that we could not test the corresponding method since no open-source implementation.}  
\begin{tabular}{|c|c|c|c|c|c|}
\hline
Methods & $L_{1} \downarrow$ & $PNSR \uparrow$ & $SSIM \uparrow$ & $LightCNN \uparrow$ & $evoLVe \uparrow$ \\
\hline 
\cite{deng2019accurate}  & 0.05 & 26.58 & 0.83 & 0.72 & 0.64 \\
\hline
\cite{gecer2019ganfit}  & / & 26.5 & 0.898 & / & / \\
\hline
\cite{jiangke2020towards}   & 0.034 & \best{29.69} & 0.89 & 0.90 & 0.85 \\
\hline
Ours\_OC  & 0.02 & 24.88 & 0.89 & 0.91 & 0.83 \\
\hline
Ours\_OD  & \best{0.01} & 28.90 & \best{0.93} & \best{0.93} & \best{0.86} \\
\hline
\end{tabular}
\label{tab:quant}
\end{table}

\begin{figure}[!tb]
\centering
  \includegraphics[width=1.0\linewidth]{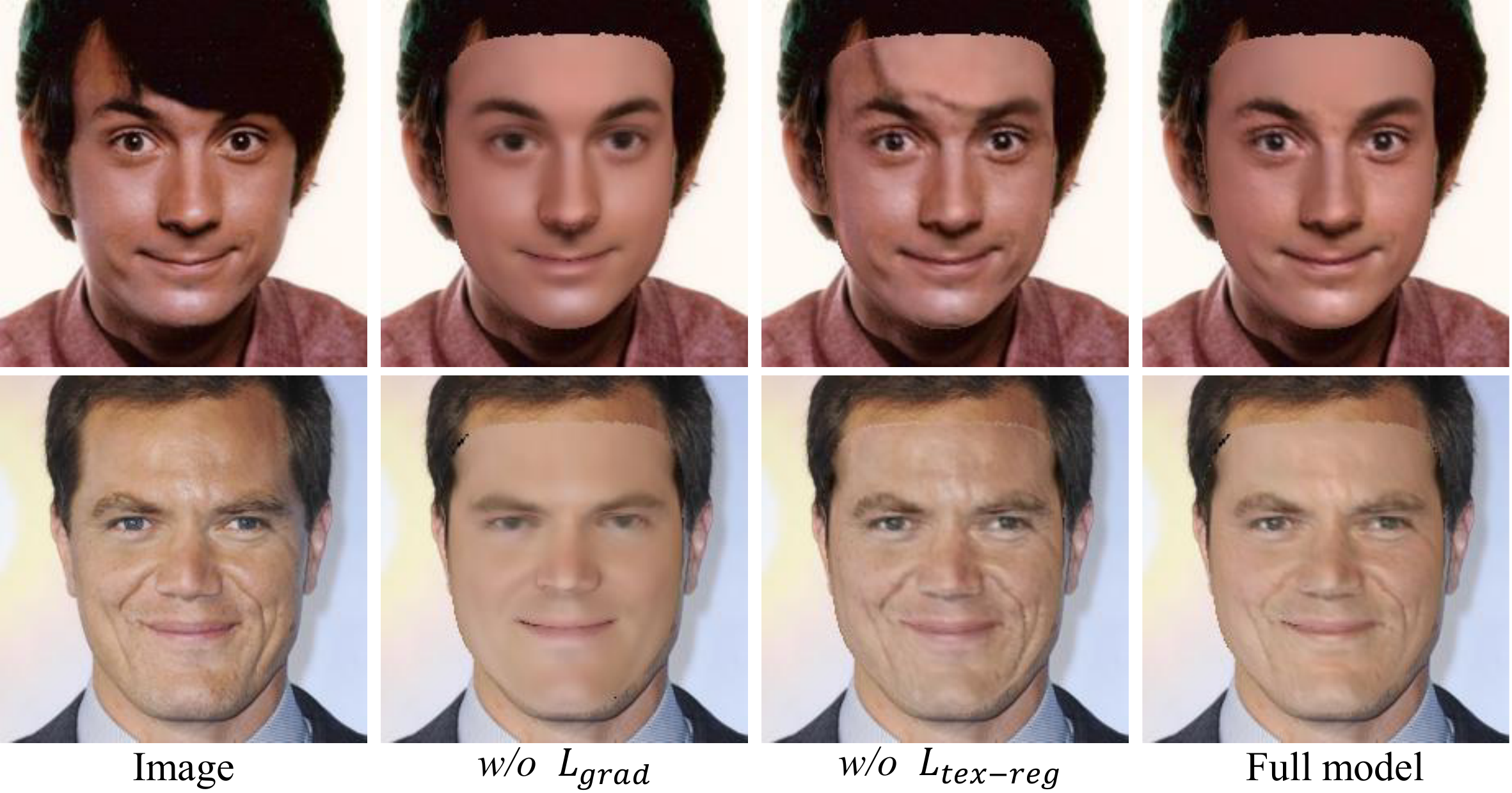}   
  \caption{\textbf{Ablation study of the proposed gradient loss and texture regularization loss:} our full model produces the most convincing results than others.}
  \label{fig:ablation}
\end{figure}

\subsection{Ablation Study}
\noindent\textbf{Effectiveness of gradient and texture regularization losses. }
We first demonstrate the functionality of the gradient loss and texture regularization loss in our pipeline using the detailed rendering results with coarse illumination and detailed albedo. As shown in Fig. \ref{fig:ablation}, our proposed $L_{grad}$ helps our model to capture the detailed information from the facial image. Our $L_{tex-reg}$ loss contributes to the disentanglement of illumination and completes the occlusion part according to the prior albedo which renders the detailed albedo map more similar to the prior albedo map. By contrast, our full model produces the most convincing results than others.


    
\begin{figure}[!tb]
\centering
  \includegraphics[width=0.95\linewidth]{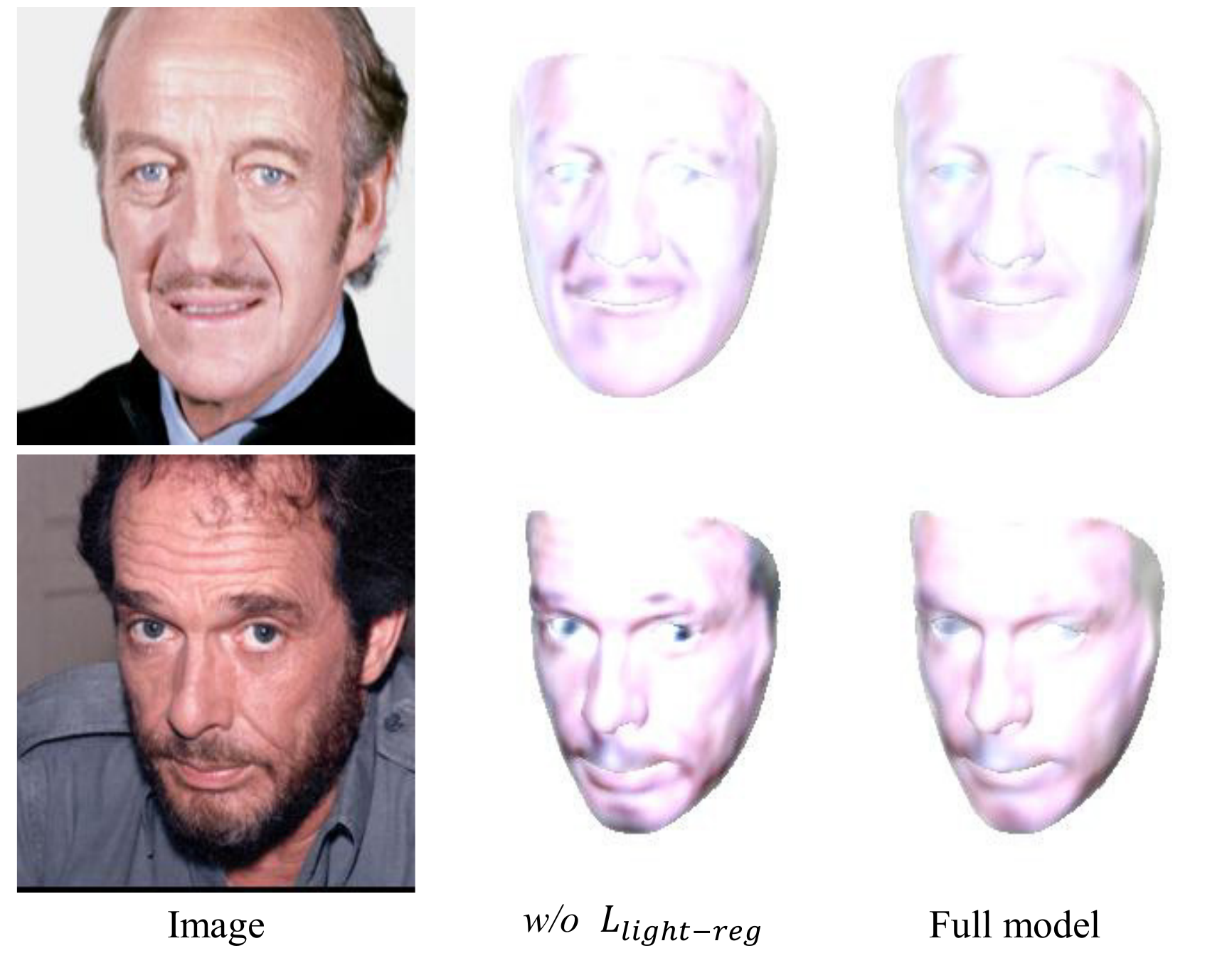}   
  \caption{\textbf{Ablation study of the proposed perceptual loss for lighting disentanglement:} the illumination image generated by our full model separates the facial intrinsic details better.}
  \label{fig:ablation_light}
\end{figure}

\noindent\textbf{Effectiveness of light regularization.}
To evaluate the effect of our light perceptual regularization loss, we perform an ablation study by showing the rendered detailed illumination images with and without this loss.
In Fig. \ref{fig:ablation_light}, the light perceptual regularization loss helps the disentanglement of illumination with facial characteristics. The illumination map recovered with the help of light perceptual regularization loss includes less facial wrinkles and beard than the one that recovered without this loss. This phenomenon indicates that the facial details are all mostly encoded in the detailed albedo map. Our illumination map has only environmental light information as far as possible thus, it is more suitable for re-renderable 3D facial generation.


    

\subsection{Limitations}
Although our model achieves competitive results on most of in-the-wild facial image datasets, it may still generate unreliable results on huge occlusion cases. The reason is that the prior model we used is only able to produce low-fidelity prior albedo and the lack of information in significantly occluded regions cannot be complemented with only symmetry regularization. Moreover, though our model takes 3DMM reconstructed albedo as prior, our reconstructed albedo is not completely independent of input image quality and generate better result when inputs are of high-resolution. Finally, the resolution of our uv map are only 256, which limits the model's capability to adapt to high-resolution input images.

    \section{Conclusion and Future Work}
\label{sec:conclusion}

We have presented a novel self-supervised neural network for 3D face reconstruction, with an emphasis on generating re-renderable high-fidelity textures from single images. We utilize the coarse 3DMM model as a prior and fine-tune on it to capture more facial details. 
We compare our results with state-of-the-art methods in qualitative and quantitative ways. The comparison demonstrates that our method does not require capturing high-resolution face texture datasets and we can generate re-renderable and realistic facial textures. 

In the future, we would like to construct a high-fidelity albedo map dataset and train a new generation model on it, that would significantly improve the fidelity. Second, we are interested in extending our model to reconstruct geometric details, because high-fidelity geometry and texture would lead to a more competitive and visually-appealing result. Finally, we wish to add more dynamics to our model and reconstruct animated facial details from a single image or video.

	\bibliographystyle{IEEEtran}
	\bibliography{3DFaceModeling} 

\begin{thebibliography}{10}
\providecommand{\url}[1]{#1}
\csname url@samestyle\endcsname
\providecommand{\newblock}{\relax}
\providecommand{\bibinfo}[2]{#2}
\providecommand{\BIBentrySTDinterwordspacing}{\spaceskip=0pt\relax}
\providecommand{\BIBentryALTinterwordstretchfactor}{4}
\providecommand{\BIBentryALTinterwordspacing}{\spaceskip=\fontdimen2\font plus
\BIBentryALTinterwordstretchfactor\fontdimen3\font minus
  \fontdimen4\font\relax}
\providecommand{\BIBforeignlanguage}[2]{{%
\expandafter\ifx\csname l@#1\endcsname\relax
\typeout{** WARNING: IEEEtran.bst: No hyphenation pattern has been}%
\typeout{** loaded for the language `#1'. Using the pattern for}%
\typeout{** the default language instead.}%
\else
\language=\csname l@#1\endcsname
\fi
#2}}
\providecommand{\BIBdecl}{\relax}
\BIBdecl

\bibitem{thies2016face2face}
J.~Thies, M.~Zollhofer, M.~Stamminger, C.~Theobalt, and M.~Nie{\ss}ner,
  ``Face2face: Real-time face capture and reenactment of rgb videos,'' in
  \emph{{IEEE} {Computer Vision and Pattern Recognition}}, 2016, pp.
  2387--2395.

\bibitem{Chen20223D}
S.-Y. Chen, Y.-K. Lai, S.~Xia, P.~Rosin, and L.~Gao, ``3d face reconstruction
  and gaze tracking in the hmd for virtual interaction `,'' \emph{IEEE Trans.
  Multimedia}, pp. 1--1, 2022.

\bibitem{lattas2020avatarme}
A.~Lattas, S.~Moschoglou, B.~Gecer, S.~Ploumpis, V.~Triantafyllou, A.~Ghosh,
  and S.~Zafeiriou, ``Avatarme: Realistically renderable 3d facial
  reconstruction,'' in \emph{{IEEE} {Computer Vision and Pattern Recognition}},
  2020, pp. 760--769.

\bibitem{Tu20213D}
X.~Tu, J.~Zhao, M.~Xie, Z.~Jiang, A.~Balamurugan, Y.~Luo, Y.~Zhao, L.~He,
  Z.~Ma, and J.~Feng, ``3d face reconstruction from a single image assisted by
  2d face images in the wild,'' \emph{IEEE Trans. Multimedia}, vol.~23, pp.
  1160--1172, 2021.

\bibitem{Li2017Multimodal}
H.~Li, J.~Sun, Z.~Xu, and L.~Chen, ``Multimodal 2d+3d facial expression
  recognition with deep fusion convolutional neural network,'' \emph{IEEE
  Trans. Multimedia}, vol.~19, no.~12, pp. 2816--2831, 2017.

\bibitem{Blanz2002A}
V.~Blanz, T.~Vetter, and A.~Rockwood, ``A morphable model for the synthesis of
  3d faces,'' \emph{ACM Trans. Graph. (Proc. {SIGGRAPH})}, pp. 187--194, 2002.

\bibitem{deng2019accurate}
Y.~Deng, J.~Yang, S.~Xu, D.~Chen, Y.~Jia, and X.~Tong, ``Accurate 3d face
  reconstruction with weakly-supervised learning: From single image to image
  set,'' in \emph{Proceedings of the IEEE Conference on Computer Vision and
  Pattern Recognition Workshops}, 2019, pp. 0--0.

\bibitem{genova2018unsupervised}
K.~Genova, F.~Cole, A.~Maschinot, A.~Sarna, D.~Vlasic, and W.~T. Freeman,
  ``Unsupervised training for 3d morphable model regression,'' in \emph{{IEEE}
  {Computer Vision and Pattern Recognition}}, 2018, pp. 8377--8386.

\bibitem{tewari2017mofa}
A.~Tewari, M.~Zollhofer, H.~Kim, P.~Garrido, F.~Bernard, P.~Perez, and
  C.~Theobalt, ``Mofa: Model-based deep convolutional face autoencoder for
  unsupervised monocular reconstruction,'' in \emph{Proceedings of the IEEE
  International Conference on Computer Vision Workshops}, 2017, pp. 1274--1283.

\bibitem{gecer2019ganfit}
B.~Gecer, S.~Ploumpis, I.~Kotsia, and S.~Zafeiriou, ``Ganfit: Generative
  adversarial network fitting for high fidelity 3d face reconstruction,'' in
  \emph{{IEEE} {Computer Vision and Pattern Recognition}}, 2019, pp.
  1155--1164.

\bibitem{tran2019towards}
L.~Tran, F.~Liu, and X.~Liu, ``Towards high-fidelity nonlinear 3d face
  morphable model,'' in \emph{{IEEE} {Computer Vision and Pattern
  Recognition}}, 2019, pp. 1126--1135.

\bibitem{tran2018nonlinear}
L.~Tran and X.~Liu, ``Nonlinear 3d face morphable model,'' in \emph{{IEEE}
  {Computer Vision and Pattern Recognition}}, 2018, pp. 7346--7355.

\bibitem{jiangke2020towards}
J.~Lin, Y.~Yuan, T.~Shao, and K.~Zhou, ``Towards high-fidelity 3d face
  reconstruction from in-the-wild images using graph convolutional networks,''
  in \emph{{IEEE} {Computer Vision and Pattern Recognition}}, 2020, pp.
  5891--5900.

\bibitem{chen2019self}
Y.~Chen, F.~Wu, Z.~Wang, Y.~Song, Y.~Ling, and L.~Bao, ``Self-supervised
  learning of detailed 3d face reconstruction,'' \emph{IEEE Trans. Image
  Process.}, pp. 8696--8705, 2020.

\bibitem{chen2019photo}
A.~Chen, Z.~Chen, G.~Zhang, K.~Mitchell, and J.~Yu, ``Photo-realistic facial
  details synthesis from single image,'' in \emph{{IEEE} {International
  Conference on Computer Vision}}, 2019, pp. 9429--9439.

\bibitem{tewari2019fml}
A.~Tewari, F.~Bernard, P.~Garrido, G.~Bharaj, M.~Elgharib, H.-P. Seidel,
  P.~Pérez, M.~Zollhöfer, and C.~Theobalt, ``Fml: Face model learning from
  videos,'' \emph{{IEEE} {Computer Vision and Pattern Recognition}}, pp.
  10\,812--10\,822, 2019.

\bibitem{tewari2018self}
A.~Tewari, M.~Zollh{\"o}fer, P.~Garrido, F.~Bernard, H.~Kim, P.~P{\'e}rez, and
  C.~Theobalt, ``Self-supervised multi-level face model learning for monocular
  reconstruction at over 250 hz,'' in \emph{{IEEE} {Computer Vision and Pattern
  Recognition}}, 2018, pp. 2549--2559.

\bibitem{jackson2017large}
A.~S. Jackson, A.~Bulat, V.~Argyriou, and G.~Tzimiropoulos, ``Large pose 3d
  face reconstruction from a single image via direct volumetric cnn
  regression,'' in \emph{{IEEE} {International Conference on Computer Vision}},
  2017, pp. 1031--1039.

\bibitem{lattas2021avatarme++}
A.~Lattas, S.~Moschoglou, S.~Ploumpis, B.~Gecer, A.~Ghosh, and S.~P. Zafeiriou,
  ``Avatarme++: Facial shape and brdf inference with photorealistic
  rendering-aware gans,'' \emph{{IEEE} Trans. Pattern Anal. Mach. Intell.},
  no.~01, pp. 1--1, 2021.

\bibitem{cao2013facewarehouse}
C.~Cao, Y.~Weng, S.~Zhou, Y.~Tong, and K.~Zhou, ``Facewarehouse: A 3d facial
  expression database for visual computing,'' \emph{IEEE Trans. Vis. Comput.
  Graph}, pp. 413--425, 2013.

\bibitem{Ferrari2017}
C.~Ferrari, G.~Lisanti, S.~Berretti, and A.~D. Bimbo, ``A dictionary
  learning-based 3d morphable shape model,'' \emph{IEEE Trans. Multimedia},
  vol.~19, no.~12, pp. 2666--2679, 2017.

\bibitem{li2017learning}
T.~Li, T.~Bolkart, J.~M. Black, H.~Li, and J.~Romero, ``Learning a model of
  facial shape and expression from 4d scans,'' \emph{ACM Trans. Graph.}, pp.
  194:1--194:17, 2017.

\bibitem{booth2018large}
J.~Booth, A.~Roussos, A.~Ponniah, D.~Dunaway, and S.~Zafeiriou, ``Large scale
  3d morphable models,'' \emph{Int. Journal of Computer Vision}, pp. 233--254,
  2018.

\bibitem{3DMM_survey}
B.~Egger, W.~A.~P. Smith, A.~Tewari, S.~Wuhrer, M.~Zollhoefer, T.~Beeler,
  F.~Bernard, T.~Bolkart, A.~Kortylewski, S.~Romdhani, C.~Theobalt, V.~Blanz,
  and T.~Vetter, ``3d morphable face models - past, present and future,''
  \emph{ACM Trans. Graph.}, 2020.

\bibitem{FLAME}
T.~Li, T.~Bolkart, M.~J. Black, H.~Li, and J.~Romero, ``Learning a model of
  facial shape and expression from {4D} scans,'' \emph{ACM Trans. Graph. (Proc.
  {SIGGRAPH Asia})}, vol.~36, no.~6, pp. 194:1--194:17, 2017.

\bibitem{karras2017progressive}
T.~Karras, T.~Aila, S.~Laine, and J.~Lehtinen, ``Progressive growing of gans
  for improved quality, stability, and variation,'' \emph{arXiv preprint
  arXiv:1710.10196}, 2017.

\bibitem{zollhofer2018state}
M.~Zollh{\"o}fer, J.~Thies, P.~Garrido, D.~Bradley, T.~Beeler, P.~P{\'e}rez,
  M.~Stamminger, M.~Nie{\ss}ner, and C.~Theobalt, ``State of the art on
  monocular 3d face reconstruction, tracking, and applications,'' in
  \emph{Comput. Graph. Forum}, vol.~37, no.~2, 2018, pp. 523--550.

\bibitem{richardson2017learning}
E.~Richardson, M.~Sela, R.~Or-El, and R.~Kimmel, ``Learning detailed face
  reconstruction from a single image,'' \emph{{IEEE} {Computer Vision and
  Pattern Recognition}}, pp. 5553--5562, 2017.

\bibitem{Fan2021Dual}
X.~Fan, S.~Cheng, K.~Huyan, M.~Hou, R.~Liu, and Z.~Luo, ``Dual neural networks
  coupling data regression with explicit priors for monocular 3d face
  reconstruction,'' \emph{IEEE Trans. Multimedia}, vol.~23, pp. 1252--1263,
  2021.

\bibitem{yang2020facescape}
H.~Yang, H.~Zhu, Y.~Wang, M.~Huang, Q.~Shen, R.~Yang, and X.~Cao, ``Facescape:
  a large-scale high quality 3d face dataset and detailed riggable 3d face
  prediction,'' in \emph{{IEEE} {Computer Vision and Pattern Recognition}},
  2020, pp. 601--610.

\bibitem{nagano2018pagan}
K.~Nagano, J.~Seo, J.~Xing, L.~Wei, Z.~Li, S.~Saito, A.~Agarwal, J.~Fursund,
  and H.~Li, ``pagan: real-time avatars using dynamic textures,'' \emph{ACM
  Trans. Graph.}, pp. 1--12, 2018.

\bibitem{yamaguchi2018high-fidelity}
S.~Yamaguchi, S.~Saito, K.~Nagano, Y.~Zhao, W.~Chen, K.~Olszewski,
  S.~Morishima, and H.~Li, ``High-fidelity facial reflectance and geometry
  inference from an unconstrained image,'' \emph{ACM Trans. Graph.}, pp.
  162:1--162:14, 2018.

\bibitem{GZCVVPT16}
P.~Garrido, M.~Zollh{\"o}fer, D.~Casas, L.~Valgaerts, K.~Varanasi, P.~Perez,
  and C.~Theobalt, ``Reconstruction of personalized 3d face rigs from monocular
  video,'' \emph{ACM Trans. Graph.}, pp. 28:1--28:15, 2016.

\bibitem{shangzhe2019unsupervised}
S.~Wu, C.~Rupprecht, and A.~Vedaldi, ``Unsupervised learning of probably
  symmetric deformable 3d objects from images in the wild,'' in \emph{{IEEE}
  {Computer Vision and Pattern Recognition}}, 2020, pp. 1--10.

\bibitem{gafni2021nerface}
G.~Gafni, J.~Thies, M.~Zollh{\"o}fer, and M.~Nie{ss}ner, ``Dynamic neural
  radiance fields for monocular 4d facial avatar reconstruction,'' in
  \emph{{IEEE} {Computer Vision and Pattern Recognition}}, 2021.

\bibitem{deferred19}
J.~Thies, M.~Zollhöfer, and M.~Nießner, ``Deferred neural rendering: image
  synthesis using neural textures,'' \emph{ACM Trans. Graph.}, pp. 1--12, 2019.

\bibitem{kim2018deep}
H.~Kim, P.~Garrido, A.~Tewari, W.~Xu, J.~Thies, M.~Niessner, P.~P{\'e}rez,
  C.~Richardt, M.~Zollh{\"o}fer, and C.~Theobalt, ``Deep video portraits,''
  \emph{ACM Trans. Graph.}, pp. 1--14, 2018.

\bibitem{decasig2021}
Y.~Feng, H.~Feng, M.~J. Black, and T.~Bolkart, ``Learning an animatable
  detailed {3D} face model from in-the-wild images,'' \emph{ACM Trans. Graph.},
  vol.~40, no.~8, 2021.

\bibitem{xueying2020lightweight}
X.~Wang, Y.~Guo, B.~Deng, and J.~Zhang, ``Lightweight photometric stereo for
  facial details recovery,'' \emph{{IEEE} {Computer Vision and Pattern
  Recognition}}, pp. 737--746, 2020.

\bibitem{deng2018uv-gan}
J.~Deng, S.~Cheng, N.~Xue, Y.~Zhou, and S.~Zafeiriou, ``Uv-gan: Adversarial
  facial uv map completion for pose-invariant face recognition,'' \emph{{IEEE}
  {Computer Vision and Pattern Recognition}}, pp. 7093--7102, 2018.

\bibitem{Chai2021Expression}
X.~Chai, J.~Chen, C.~Liang, D.~Xu, and C.-W. Lin, ``Expression-aware face
  reconstruction via a dual-stream network,'' \emph{IEEE Trans. Multimedia},
  vol.~23, pp. 2998--3012, 2021.

\bibitem{zhang2006face}
L.~Zhang and D.~Samaras, ``Face recognition from a single training image under
  arbitrary unknown lighting using spherical harmonics,'' \emph{{IEEE} Trans.
  Pattern Anal. Mach. Intell.}, pp. 351--363, 2006.

\bibitem{paysan2009a}
P.~Paysan, R.~Knothe, B.~Amberg, S.~Romdhani, and T.~Vetter, ``A 3d face model
  for pose and illumination invariant face recognition,'' \emph{2009 Sixth IEEE
  International Conference on Advanced Video and Signal Based Surveillance},
  pp. 296--301, 2009.

\bibitem{deng2018arcface}
J.~Deng, J.~Guo, X.~Niannan, and S.~Zafeiriou, ``Arcface: Additive angular
  margin loss for deep face recognition,'' in \emph{{IEEE} {Computer Vision and
  Pattern Recognition}}, 2019, pp. 4690--4699.

\bibitem{pix2pix2017}
P.~Isola, J.-Y. Zhu, T.~Zhou, and A.~A. Efros, ``Image-to-image translation
  with conditional adversarial networks,'' \emph{{IEEE} {Computer Vision and
  Pattern Recognition}}, pp. 1125--1134, 2017.

\bibitem{yu2018bisenet}
C.~Yu, J.~Wang, C.~Peng, C.~Gao, G.~Yu, and N.~Sang, ``Bisenet: Bilateral
  segmentation network for real-time semantic segmentation,'' \emph{{European
  Conference on Computer Vision}}, pp. 325--341, 2018.

\bibitem{liu2015faceattributes}
Z.~Liu, P.~Luo, X.~Wang, and X.~Tang, ``Deep learning face attributes in the
  wild,'' in \emph{{IEEE} {International Conference on Computer Vision}}, 2015,
  pp. 3730--3738.

\bibitem{feng2018joint}
Y.~Feng, F.~Wu, X.~Shao, Y.~Wang, and X.~Zhou, ``Joint 3d face reconstruction
  and dense alignment with position map regression network,'' in
  \emph{{European Conference on Computer Vision}}, 2018, pp. 534--551.

\bibitem{he2016deep}
K.~He, X.~Zhang, S.~Ren, and J.~Sun, ``Deep residual learning for image
  recognition,'' in \emph{{IEEE} {Computer Vision and Pattern Recognition}},
  2016, pp. 770--778.

\bibitem{zhu2016face}
X.~Zhu, Z.~Lei, X.~Liu, H.~Shi, and S.~Z. Li, ``Face alignment across large
  poses: A 3d solution,'' in \emph{{IEEE} {Computer Vision and Pattern
  Recognition}}, 2016, pp. 146--155.

\bibitem{shi2020neutral}
T.~Shi, Z.~Zou, X.~Song, Z.~Song, C.~Gu, C.~Fan, and Y.~Yuan, ``Neutral face
  game character auto-creation via pokerface-gan,'' in \emph{ACM International
  Conference on Multimedia}, 2020, pp. 3201--3209.

\bibitem{CelebAMask-HQ}
C.-H. Lee, Z.~Liu, L.~Wu, and P.~Luo, ``Maskgan: Towards diverse and
  interactive facial image manipulation,'' in \emph{{IEEE} {Computer Vision and
  Pattern Recognition}}, 2020, pp. 5549--5558.

\bibitem{schroff2015facenet}
F.~Schroff, D.~Kalenichenko, and J.~Philbin, ``Facenet: A unified embedding for
  face recognition and clustering,'' in \emph{{IEEE} {Computer Vision and
  Pattern Recognition}}, 2015, pp. 815--823.

\bibitem{schneider2017efficient}
A.~Schneider, S.~Schonborn, L.~Frobeen, B.~Egger, and T.~Vetter, ``Efficient
  global illumination for morphable models,'' in \emph{{IEEE} {International
  Conference on Computer Vision}}, 2017, pp. 3865--3873.

\bibitem{wu2018light}
X.~Wu, R.~He, Z.~Sun, and T.~Tan, ``A light cnn for deep face representation
  with noisy labels,'' \emph{IEEE Transactions on Information Forensics and
  Security}, pp. 2884--2896, 2018.

\bibitem{zhao2019multi}
J.~Zhao, J.~Li, X.~Tu, F.~Zhao, Y.~Xin, J.~Xing, H.~Liu, S.~Yan, and J.~Feng,
  ``Multi-prototype networks for unconstrained set-based face recognition,'' in
  \emph{IJCAI}, 2019.

\end{thebibliography}

\end{document}